\documentclass[lettersize,journal,twoside]{IEEEtran}
\usepackage{amsmath, amsfonts, amsthm}
\usepackage{array}
\usepackage[caption=false,font=normalsize,labelfont=sf,textfont=sf]{subfig}
\usepackage{textcomp}
\usepackage{stfloats}
\usepackage{url}
\usepackage{verbatim}
\usepackage{graphicx}
\usepackage{cite}
\usepackage{soul}
\usepackage{xcolor}
\usepackage{tabularx}
\newcolumntype{Y}{>{\centering\arraybackslash}X}
\usepackage{booktabs}
\usepackage{hyperref}
\usepackage{multicol}
\usepackage{multirow}
\usepackage{algorithm}
\usepackage{algpseudocode}
\usepackage{scalerel}
\usepackage{tikz}
\usetikzlibrary{svg.path}

\newtheorem{definition}{Definition}
\newtheorem{theorem}{Theorem}
\newtheorem{corollary}{Corollary}

\begin{document}

\title{PHLP: Sole Persistent Homology for Link Prediction - Interpretable Feature Extraction}

\author{
Junwon~You, 
Eunwoo~Heo,
Jae-Hun~Jung
\thanks{\textit{Junwon You and Eunwoo Heo are co-first authors.}}
\thanks{}
\thanks{Junwon You, Eunwoo Heo, and J.-H. Jung are with the Department of Mathematics, Pohang University of Science and Technology (POSTECH), Pohang, South Korea. 
J.-H. Jung is also with the Graduate School of Artificial Intelligence, POSTECH, Pohang, South Korea.}
\thanks{}
\thanks{Preprint. Under review.}}

\maketitle
\begin{abstract}
Link prediction (LP), inferring the connectivity between nodes, is a significant research area in graph data, where a link represents essential information on relationships between nodes. 
Although graph neural network (GNN)-based models have achieved high performance in LP, understanding why they perform well is challenging because most comprise complex neural networks. 
We employ persistent homology (PH), a topological data analysis method that helps analyze the topological information of graphs, to interpret the features used for prediction.
We propose a novel method that employs PH for LP (PHLP) focusing on how the presence or absence of target links influences the overall topology. 
The PHLP utilizes the \textit{angle hop subgraph} and new node labeling called \textit{degree double radius node labeling (Degree DRNL)}, distinguishing the information of graphs better than DRNL. Using only a classifier, PHLP performs similarly to state-of-the-art (SOTA) models on most benchmark datasets. 
Incorporating the outputs calculated using PHLP into the existing GNN-based SOTA models improves performance across all benchmark datasets. 
To the best of our knowledge, PHLP is the first method of applying PH to LP without GNNs. 
The proposed approach, employing PH while not relying on neural networks, enables the identification of crucial factors for improving performance.
\end{abstract}

\section{Introduction}

Graph data pervade numerous domains such as social networks, biological systems, recommendation engines, and e-commerce networks~\cite{zhang2020deep, wu2020comprehensive}.
The graph is well-suited for modeling complex real-world relationships.

Predicting missing or potential connections within a graph is essential for many applications, unlocking valuable insight and facilitating intelligent decision-making. 
The ability to predict future network interactions can be applied to diverse domains, including friend recommendations on social networks~\cite{adamic2003friends, yao2016link, fire2011link}, knowledge graph completion~\cite{kazemi2018simple, nayyeri2021link}, identification of potential drug-protein interactions in bioinformatics~\cite{stanfield2017drug, nasiri2021novel}, prediction protein interactions~\cite{lei2013novel, kovacs2019network, nasiri2021novel}, and optimization of supply chain logistics~\cite{brockmann2022supply, brintrup2018predicting}.

The link prediction (LP) problem has been categorized into three major paradigms: heuristic methods, embedding methods, and graph neural network (GNN)-based methods, which are explored in detail in Section~\ref{sec:related works}.
Recently, compared to heuristic~\cite{adamic2003friends, lu2011link, barabasi1999emergence, zhou2009predicting, brin2012reprint, jeh2002simrank} and embedding methods~\cite{koren2009matrix, perozzi2014deepwalk, grover2016node2vec, tang2015line}, GNN-based models have achieved significant score improvements in capturing intricate relationships within graphs~\cite{kipf2016variational, zhang2018link, yun2021neo, mavromatis2020graph, yan2021link, pan2021neural}. 

\begin{figure}[!htbp]
\centering
\includegraphics[width=\linewidth]{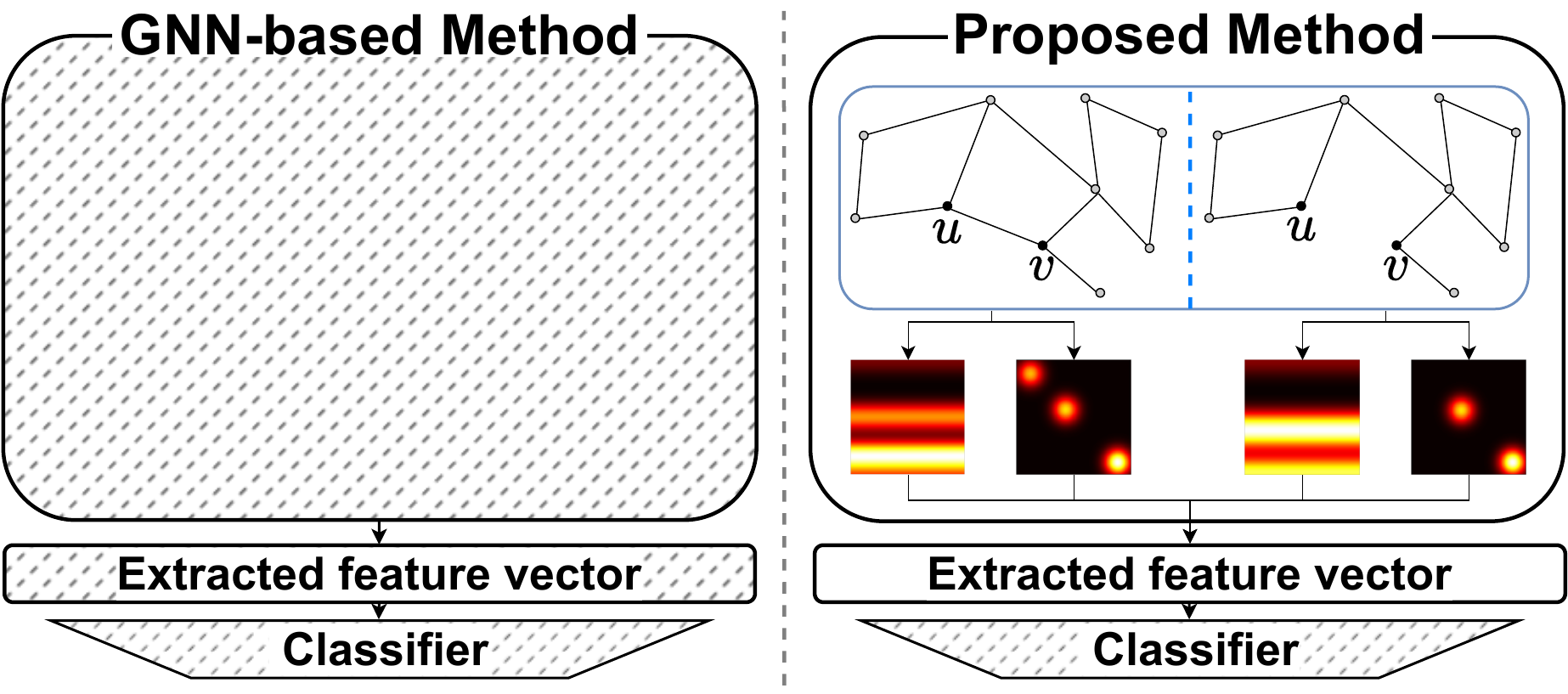} 
\caption{Difference between the GNN-based and proposed methods.
(Left) The GNN-based method extracts feature vectors through optimization (dashed area), making it difficult to interpret what these vectors represent. (Right) The proposed method extracts feature vectors through the designed analysis process, resulting in interpretable vectors.}
\label{fig:Blackbox}
\end{figure}

However, GNN-based methods are comprised of neural networks, making it challenging to understand the reasons for their performance.
To explore these reasons, we employ persistent homology (PH), a mathematical tool in topological data analysis (TDA) that enables the inference of topological information regarding the manifold approximating the data~\cite{huber2021persistent, dey2022computational} by quantifying the persistence of topological features across multiple scales.
Various research has had successful outcomes in applying PH to graph classification and node classification tasks~\cite{horn2021topological, ye2023treph, carriere2020perslay, taiwo2024explaining, wen2024tensor, immonen2024going, ying2024boosting, zhao2019learning, chen2021topological, zhao2020persistence}.
In contrast, relatively few studies have explored using PH for LP.
The topological loop-counting (TLC) GNN~\cite{yan2021link} is a notable example that uses PH.
The TLC-GNN injects topological information into a GNN, and experiments were conducted on benchmark data where node attributes are available.

\begin{figure*}[!htbp]
\centering
\includegraphics[width=\linewidth]{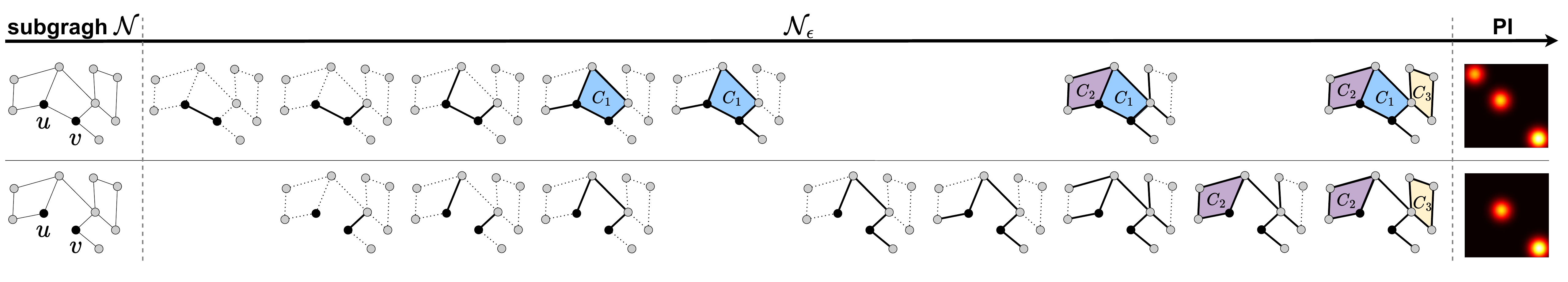}
\vspace{-7mm}
\caption{Topological features in subgraphs with and without a target link $(u,v)$.
The diagram illustrates the topological information extraction process for the subgraph $\mathcal{N}$, as described in Section~\ref{subsec:persistenthomology}. 
The presence (top) or absence (bottom) of the target link changes the topological structure of the graph. 
Top row: When the target link is connected, three features ($C_1$, $C_2$, and $C_3$) are detected shown in the persistence image (PI) in the right column.
The PI represents the topological features of the subgraph $\mathcal{N}$ (Section~\ref{subsec:persistenceimage}).
Bottom row: When the target link is absent, only two features ($C_2$ and $C_3$) are detected as depicted in the corresponding PI. }
\label{motivation}
\end{figure*}

In this context, as illustrated in Figure~\ref{fig:Blackbox}, we present a novel approach to LP, called PHLP, which calculates the topological information of a graph.
The main difference between GNN-based method and our method is described in Section~\ref{subsec:interpretability}.
To use the topological information of subgraphs for LP, we measure how the topological information changes depending on the existence of the target link, as illustrated in Figure~\ref{motivation}.
To extract topological information from various perspectives, we utilize \textit{angle hop subgraphs} for each target node. 
Additionally, we propose new node labeling called \textit{degree double radius node labeling (Degree DRNL)}, which incorporates degree information for each node, using DRNL~\cite{zhang2018link}.

The contributions are summarized as follows:
\begin{itemize}
    \item We develop an explainable LP method, PHLP, that employs the topological information for LP through PH without relying on neural networks, as illustrated in Figure~\ref{fig:Blackbox}.
    \item We demonstrate that the proposed method, even with a simple classifier such as a multilayer perceptron (MLP), can achieve LP performance close to that of state-of-the-art (SOTA) models.
    This method surpassed the SOTA performance for the Power dataset.
    \item We reveal that merely incorporating vectors computed by PHLP into existing LP models, including SOTA models, can improve their performance.
    \item To the best of our knowledge, the proposed method using PH without a GNN is the first to achieve performance close to that of SOTA models.
\end{itemize}
\section{Related Work}\label{sec:related works}

\subsection{Link Prediction}

\noindent\textbf{Heuristic Methods.}
Heuristic-based approaches to LP compute the predefined structural features within the observed nodes and edges of the graph. 
Classic methods, such as common neighbors~\cite{adamic2003friends}, Adamic-Adar~\cite{adamic2003friends}, Jaccard coefficient~\cite{lu2011link}, and preferential attachment~\cite{barabasi1999emergence}, rely on simple heuristics that capture certain aspects of node relationships. 
Zhou \textit{et al.}~\cite{zhou2009predicting} proposed a local random walk method, whereas Jeh and Widom~\cite{jeh2002simrank} developed SimRank to quantify similarity based on the structural context. 
Although heuristic methods provide a preliminary understanding of LP, they are limited by their inability to capture complex relationships within graphs.
Furthermore, heuristic methods are effective only when the defined heuristics align with the graph structure; therefore, applying heuristic methods across all graph datasets can be challenging.

\noindent\textbf{Embedding Methods.}
Embedding methods map nodes from the graph into a low-dimensional vector space where geometric relationships mirror the graph structure. 
Koren \textit{et al.}~\cite{koren2009matrix} demonstrated the power of matrix factorization for collaborative filtering. 
Perozzi \textit{et al.}~\cite{perozzi2014deepwalk} introduced DeepWalk, using random walks to generate node sequences and employing the skip-gram model to produce embeddings. 
Tang \textit{et al.}~\cite{tang2015line} developed large-scale information network embedding (LINE), which preserves local and global structures. 
Grover and Leskovec~\cite{grover2016node2vec} further advanced this approach with Node2Vec (N2V), proposing a flexible notion of the neighborhood to capture diverse node relationships.

\begin{figure*}[!htbp]
\centering
\includegraphics[width=18cm]{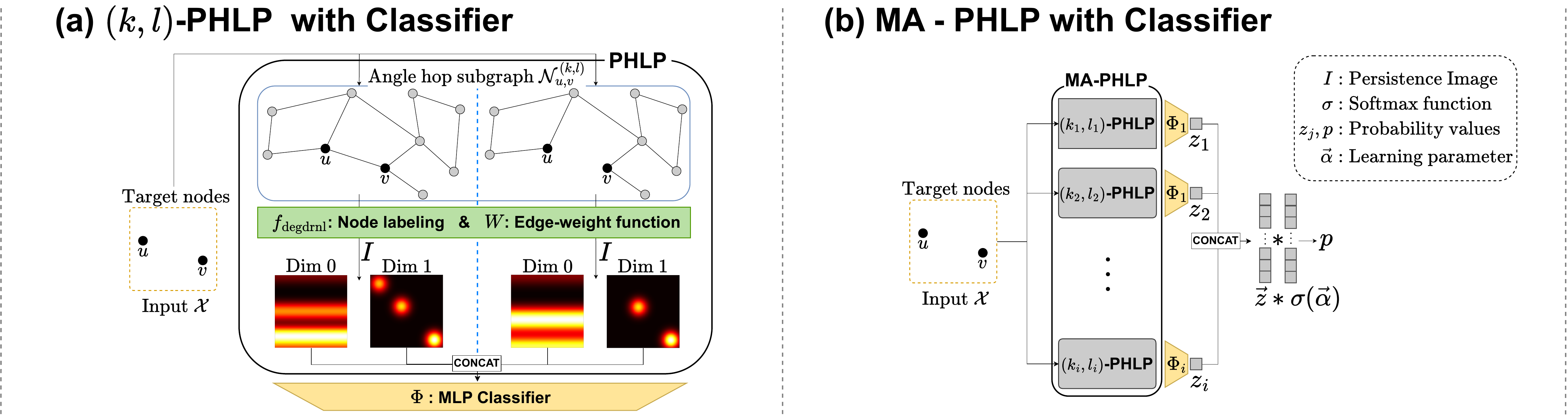}
\caption{Overall structure of persistent homology for link prediction (PHLP) and multiangle PHLP (MA-PHLP). (a) PHLP calculates the topological information based on the existence of target links in angle hop subgraphs for each target node. (b) With a classifier, MA-PHLP integrates topological information across various angles to perform LP. }
\label{fig:pipeline}
\end{figure*}

Embedding methods are advantageous due to their applicability regardless of the data characteristics using optimization. Node representations capture global properties and long-range effects through the learning process. However, these methods often require significantly large dimensions to express basic heuristics, resulting in lower performance than heuristic methods~\cite{nickel2014reducing}.
Moreover, in embedding methods, Ribeiro \textit{et al.}~\cite{ribeiro2017struc2vec} explained that two nodes with similar neighborhood structures may have vastly different embedded vectors, especially when they are far apart in the graph, leading to incorrect predictions.

\noindent\textbf{GNN-Based Methods.}
The GNN has become a pivotal approach to LP due to its ability to grasp graph-structured data. 
By effectively incorporating local and global information through message passing and graph aggregation layers, GNNs enhance LP performance.
The model by Zhang \textit{et el.}~\cite{zhang2018link} uses subgraphs as the primary structural units to learn and predict connections, resulting in significant improvement. 
This paradigm shift led to research focusing on refining and advancing subgraph methods in the context of GNNs~\cite{yun2021neo, mavromatis2020graph, pan2021neural, cai2022line, chamberlain2023graph, zhang2024transformative, wang2024a, li2024subgraph, nie2025local}. 
However, despite their superior performance, GNN-based methods pose a challenge in comprehending the underlying mechanisms driving their predictions.
Within this context,
we develop the PHLP, based on PH, with performance comparable to GNN-based models. 

\subsection{Persistent Homology on Graph Data}
In recent years, PH, a method of analyzing the topological features of data, has been widely used to analyze graph data. 
It has demonstrated its effectiveness in graph classification tasks by analyzing the topology of graphs~\cite{horn2021topological, ye2023treph, carriere2020perslay, taiwo2024explaining, wen2024tensor, immonen2024going, ying2024boosting, zhao2019learning} and has been applied to node classification tasks~\cite{horn2021topological, chen2021topological, zhao2020persistence}. 
Bhatia \textit{et al.}~\cite{bhatia2019persistent} successfully proposed applying PH to LP problem.
However, its suitability for LP tasks has been limited, and research on applying PH for LP has progressed slowly.
Yan \textit{et al.}~\cite{yan2021link} proposed an intriguing approach by integrating PH with GNNs. 
While their model demonstrates the potential of PH for capturing topological features of graph data, it relies on GNN structures.
Additionally, the TLC-GNN requires further research on datasets without node attributes.

Although PH has demonstrated success in graph and node classification tasks, its filtration technique, tailored to analyzing the entire graph structure, might not be optimal for LP as the role of each node in LP differs from that in graph or node classification tasks. 
To address this challenge and advance research in LP, we develop a filtration method tailored explicitly to LP tasks.
\section{Method}
\subsection{Outline of the Proposed Methods}

We propose (a) PHLP and (b) multiangle PHLP (MA-PHLP) as described in Figure~\ref{fig:pipeline}. 
The PHLP method analyzes the topological structure of the graph, focusing on target links.
First, PHLP samples a $(k,l)$-angle hop subgraph for the given target nodes (Section~\ref{subsec:anglehop}). 
Then, PHLP computes persistence images (PIs; Section~\ref{subsec:persistenceimage}) for cases with and without the target link. 
To calculate PIs, we introduce the node labeling and define the edge-weight function (Section~\ref{subsec:filtration}).
Through PHLP, each target node is transformed into a vector comprising PIs.
In addition, LP is performed using the calculated vectors
with a classifier (Section~\ref{subsec:mlp}). 
To reflect diverse topological information, we also propose MA-PHLP, which analyzes data from various angles (Section~\ref{subsec:maphlp}).

\begin{figure}[!htbp]
\centering
\includegraphics[width=\linewidth]{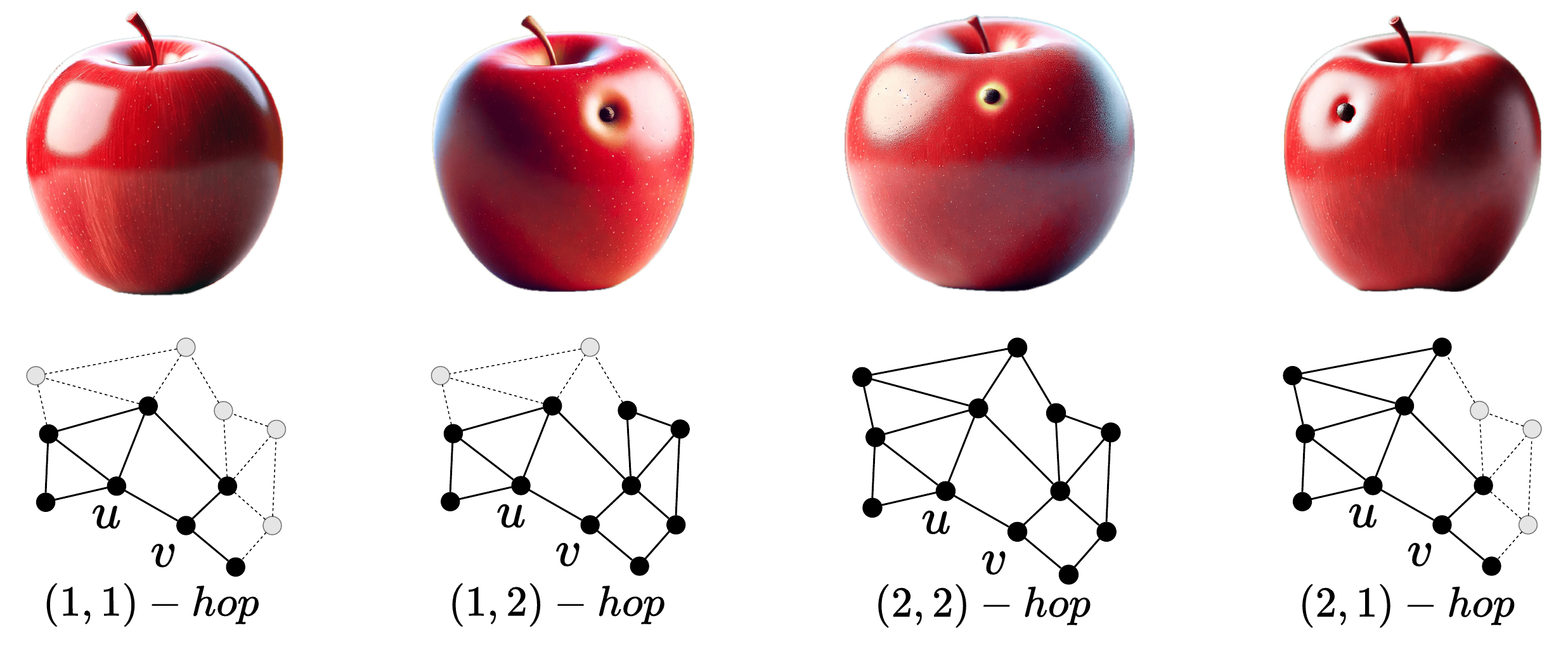} 
\caption{The motivation of $(k,l)$-angle hop subgraph. Just as viewing photographs of an apple from multiple angles provides a comprehensive understanding. This figure illustrates the capability to extract subgraphs from various perspectives.}
\label{fig:multi-angle}
\end{figure}

\subsection{Extracting Angle Hop Subgraph}
\label{subsec:anglehop}

Given a graph \( G = (V, E) \) and two nodes \( u,v \in V \), a \(k\)-hop enclosing subgraph for $(u,v)$ is defined as \( \mathcal{N}^k_{u,v} = (V', E') \) such that 

\begin{align*}
    V' &= \{ z \in V \mid d(u, z) \leq k \text{ or } d(z, v) \leq k\}, \\
    E' &= \{ (z, w) \in E \mid z \in V' \text{ and } w \in V' \},
\end{align*}
where $d(z, w)$ is the minimum number of edges in any path from $z$ to $w$ in $G$.
We define a \((k,l)\)-angle hop enclosing subgraph, where the term ``angle'' signifies viewing the subgraph from multiple perspectives. 
When defining the \((k,l)\)-angle hop subgraph, we were motivated by the observation in Figure~\ref{fig:multi-angle} that the information captured in a photograph varies depending on the angle from which it is taken. 
To confirm features like scratches on an apple, photographs from multiple angles are needed for a comprehensive understanding.
Similarly, we hypothesized that a multi-perspective approach is necessary to predict the connections between nodes \( u \) and \( v \) in a graph. 
Therefore, we devised a method to extract subgraphs by varying the degrees of separation between \( u \) and \( v \), enabling us to capture different views of the graph.

Given a graph \( G = (V, E) \) and two nodes \( u, v \in V \), a \((k,l)\)-angle hop enclosing subgraph for \( (u, v) \) is defined as \( \mathcal{N}^{(k,l)}_{u,v} = (V', E') \) such that 
\begin{align*}
    V' &= \{ z \in V \mid d(u, z) \leq k \text{ or } d(z, v) \leq l \}, \\
    E' &= \{ (z, w) \in E \mid z \in V' \text{ and } w \in V' \}.
\end{align*}
Thus, the angle hop can generate subgraphs in various forms, providing flexibility to adapt to various graph characteristics.
The variation in prediction due to angle is discussed in Section~\ref{subsec:Ablation Study}.

\subsection{Filtration of the Subgraph}
\label{subsec:filtration}

For a given subgraph, we construct the filtration to calculate the topology using PH.
Existing filtration methods for graphs do not differentiate the target link, which is crucial for predicting the presence of target link.
To extract topological features from graphs for this purpose, the filtration process must distinguish the target link.
To address this challenge, we utilize a node labeling that marks nodes based on their relative positions to the target nodes, emphasizing the importance of the target link within the graph's topology. Based on this labeling, we define an edge-weight function that is subsequently used to construct the filtration.

\begin{figure}[!htbp]
\centering
\captionsetup[subfloat]{font=small}
\subfloat[\small{DRNL}]{\includegraphics[width=0.9\linewidth]{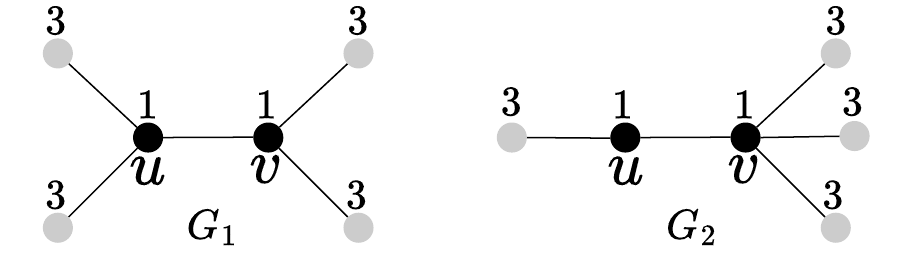}
\label{drnl}}
\hfil
\subfloat[\small{Degree DRNL}]{\includegraphics[width=0.9\linewidth]{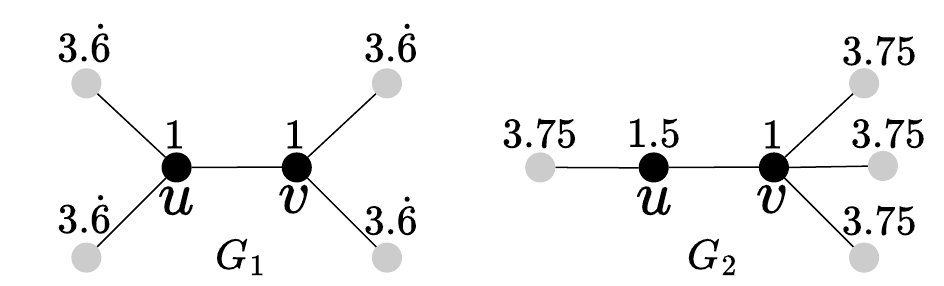}
\label{degdrnl}}
\caption{Node labeling on graphs. (a) Node label values without considering the graph structure cannot distinguish between $G_1$ and $G_2$ using DRNL. (b) Applying Degree DRNL allows $G_1$ and $G_2$ to be distinguished solely by node label values.
}
\label{nodelabel}
\end{figure}

\begin{figure}[!htbp]
\centering
\captionsetup[subfloat]{font=small}
\subfloat[\small{DRNL}]{\includegraphics[width=0.48\linewidth]{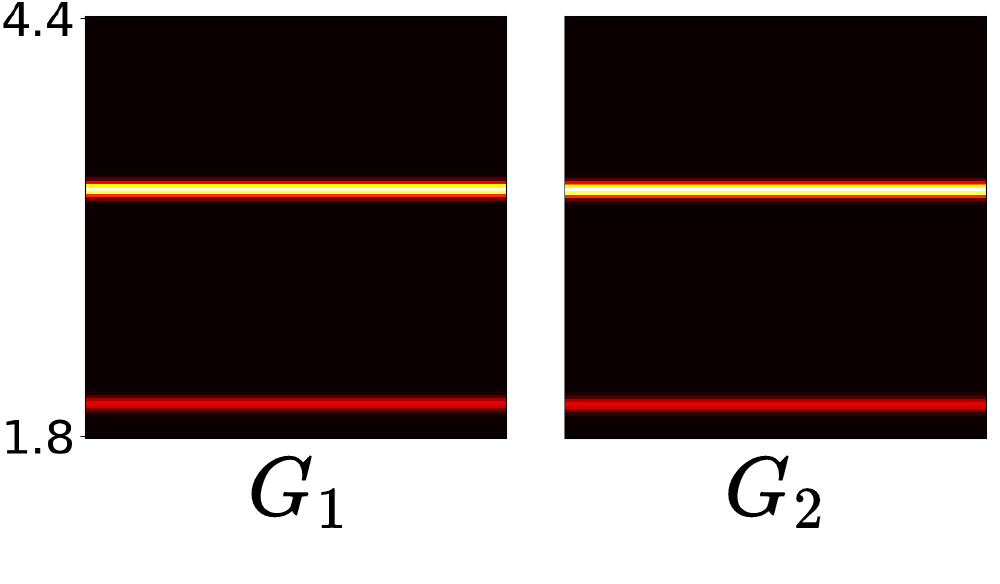}
\label{drnl_PD}}
\hfil
\subfloat[\small{Degree DRNL}]{\includegraphics[width=0.48\linewidth]{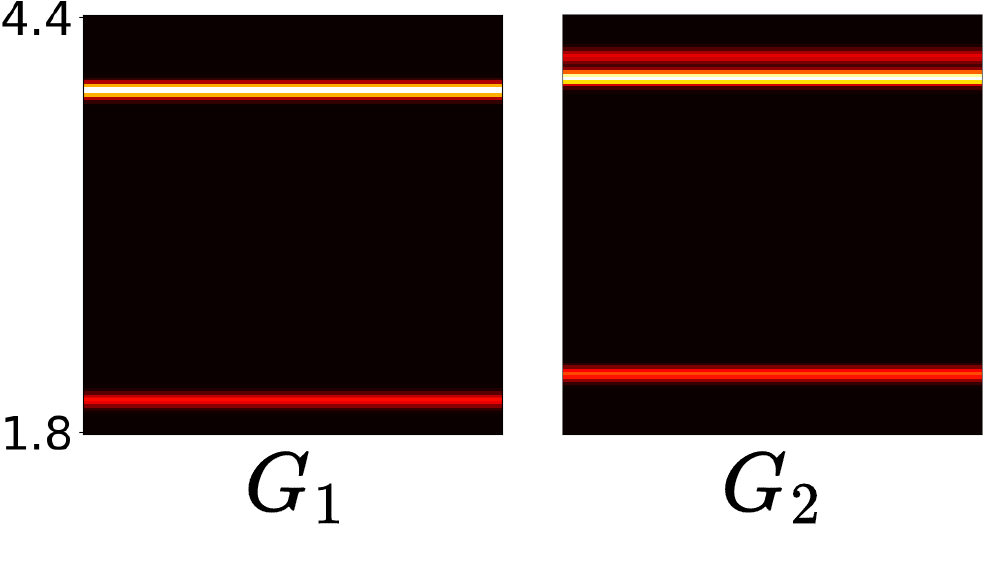}
\label{degdrnl_PD}}
\caption{Persistence images (PIs) for two node labeling methods for the graphs in Figure~\ref{nodelabel}.
(a) DRNL exhibits identical zero-dimensional PIs for $G_1$ and $G_2$, (b) Degree DRNL produces distinct outcomes, effectively distinguishing between the two.
}
\label{PD_nodelabel}
\end{figure}

\noindent\textbf{Degree DRNL.}
Zhang \textit{et al.}~\cite{zhang2018link} introduced DRNL, which computes the distance from any node to two fixed nodes.
For any subgraph $\mathcal{N}=(V',E')$ of $G$ and two nodes $a,b \in V'$, the DRNL $f^{(a,b)}_{\text{drnl}}: V' \rightarrow \mathbb{N}$ based on $(a,b)$ of $G$ for any vertex $w$ in $V'$, is defined as
\begin{equation}
    f^{(a,b)}_{\text{drnl}}(w) = 1 + \min(d(w,a), d(w,b)) + q_w(q_w + r_w - 1),
\end{equation}
where $q_w \in \mathbb{Z}$ and $r_w \in \{ 0, 1\}$ are integers representing the quotient and remainder, respectively, such that $d(w,a)+d(w,b) = 2q_w + r_w$. 
We call these two nodes, $a$ and $b$, \textit{center nodes}.
These center nodes do not need to be the target nodes used when extracting the subgraph.

However, DRNL encounters limitations when the graph is transformed into node-label information.
As depicted in Figure~\ref{drnl}, DRNL assigns the same node labels to different graphs, resulting in identical zero-dimensional PIs (Figure~\ref{drnl_PD}, Section~\ref{subsec:persistenceimage}).
To incorporate the local topology of each node with the effects of DRNL, we introduced \textit{Degree DRNL}. 
For a given subgraph \( \mathcal{N} = (V', E') \) of $G$ and center nodes $a,b \in V'$, the Degree DRNL $f^{(a,b)}_{\text{degdrnl}} : V' \rightarrow \mathbb{R}$  based on $(a,b)$, for all vertex \( w \) in \( V' \), is defined as
\begin{equation}
    f^{(a,b)}_{\text{degdrnl}}(w) = f^{(a,b)}_{\text{drnl}}(w) + \frac{M - \deg(w)}{M},
    \label{eq:degdrnl}
\end{equation}
where \( M \) denotes the maximum degree of nodes in \( \mathcal{N}\).
The $(M-\deg(w))/M$ term above assigns larger values for lower degrees of $w$. When $M = \deg(w)$, the value of Degree DRNL matches the original DRNL, ensuring that the edges connected to nodes with higher degrees are assigned smaller values, promoting their earlier emergence in the filtration.
Figure~\ref{degdrnl} demonstrates various node labels obtained using Degree DRNL, resulting in PIs that can be distinguished from each other (Figure~\ref{degdrnl_PD}). 

\noindent\textbf{Edge-weight function.} 
For a given subgraph \( \mathcal{N} = (V', E') \), \( f : V' \rightarrow \mathbb{R} \) denotes any node labeling function.
The edge-weight function $W(f):E' \rightarrow \mathbb{R}$, for any edge \( (w, z) \) in \( E' \), is defined as 
\(
W(f)(w,z) =\max(f(w),f(z)).
\)
\\

\subsection{Persistent Homology}
\label{subsec:persistenthomology}

Given an edge-weighted subgraph \( \mathcal{N} = (V', E', W) \), we construct a Rips filtration~\cite{vietoris1927hoheren, gromov1987hyperbolic, edelsbrunner2002topological} and compute its PH.
First, we create a sequence of subgraphs \( \{\mathcal{N}_{\epsilon}\}_{\epsilon \in \mathbb{R}} \), where each \( \mathcal{N}_{\epsilon} = (V', E'_\epsilon) \) and \( E'_{\epsilon}= \{ e \in E \mid W(e) \leq \epsilon \} \). 
Second, we convert each subgraph \( \mathcal{N}_{\epsilon} \) into the Rips complex 
\( K_{\epsilon} = \{ \tau \in \mathbb{X} \mid (w,z) \in E'_{\epsilon} \text{ for any two vertices } w,z \in \tau \}\), where $\mathbb{X}$ is the power set of $V'$. 
In \(K_{\epsilon}\), a simplex $\tau$ is formed when the vertices in \( \tau \) are pairwise connected by edges in \( \mathcal{N}_{\epsilon} \).
Then, the Rips filtration is obtained as  
\(
    K_{\epsilon_1} \hookrightarrow K_{\epsilon_2} \hookrightarrow \cdots \hookrightarrow K_{\epsilon_m} = \mathbb{X}
\)
for $\epsilon_1 \le \epsilon_2 \le \cdots \le \epsilon_m $.
Third, we compute the $p$-dimensional homology group \( H_p(K_{\epsilon}) \) for each complex \( K_{\epsilon}\) and track how these groups change as \( \epsilon \) increases. 
The persistence diagram $D$~\cite{edelsbrunner2002topological} comprises
persistence pairs \((b,d)\) representing the \(\epsilon\) values at which a homological feature appears $b$ and disappears $d$, respectively, in the filtration.

\subsection{Theoretical Analysis of the Proposed Filtration Method}
We construct filtrations of graphs to compute persistence diagrams and extract topological features. Existing filtrations do not explicitly consider the target link in a graph. 
We proposed a novel filtration method as explained in Section~\ref{subsec:persistenthomology} based on node labeling.
The proposed construction demonstrates stability for node labeling functions, providing a strong theoretical foundation for validity.
\begin{definition}[\textbf{Graph isomorphism}]
    A \textit{graph isomorphism} between two graphs $G_1=(V_1,E_1)$ and $G_2=(V_2,E_2)$ is a bijection between the vertex sets $f:V_1 \rightarrow V_2$ such that any two vertices $u$ and $v$ of $V_1$ are adjacent in $G_1$ if and only if $f(u)$ and $f(v)$ are adjacent in $G_2$.
\end{definition}

When representing graphs as vectors, the goal is to ensure that the same vector represents isomorphic graphs. However, in the context of the LP problem, two isomorphic graphs $G_1$ and $G_2$ can be represented by different vectors, depending on the placement of the target node. 
To address this, we redefine graph isomorphism to account for target nodes explicitly.
Originally, the target node is needed for calculating the node label, but we denote the target nodes \( (u, v) \) in graph \( G \) as \( G^{(u,v)} \).
\begin{definition}[\textbf{Graph isomorphism with target nodes}]
    A \textit{graph isomorphism with target nodes} between two graphs $G_1^{(u_1,v_1)}$ and $G_2^{(u_2,v_2)}$ is a graph isomorphism $f$ between $G_1$ and $G_2$ such that \( f(\{u_1,u_2\}) = \{v_1, v_2\} \)
\end{definition}

To guarantee the robustness of our approach under minor perturbations, we present a stability theorem that quantifies the impact of changes in node labeling on the persistence diagrams. 
\begin{theorem}[\textbf{Stability theorem}]\label{thm:stability}
    Let $G^{(u,v)} = (V,E)$ be a graph with target nodes $u,v$ in $V$ and let $f^{(u,v)}$ and $g^{(u,v)}$ be two node labeling fucntions defined on $G^{(u,v)}$. Denote the $p$-dimensional persistence diagrams of $G^{(u,v)}$ obtained from the filtrations constructed by $f^{(u,v)}$ and $g^{(u,v)}$ as $dgm_p(f^{(u,v)})$ and $dgm_p(g^{(u,v)})$, respectively. Then, the following stability inequality holds: 
    $$D_B(dgm_p(f^{(u,v)}),dgm_p(g^{(u,v)})) \leq \lVert f^{(u,v)}-g^{(u,v)} \rVert_\infty,$$
    where $D_B$ is bottleneck distance, $\lVert \cdot \rVert_\infty$ is infinity norm.
\end{theorem}

The proof of the theorem is in Appendix~\ref{sec:proof}.
This theorem aligns with the general framework of stability results for persistence diagrams~\cite{cohen2005stability}, as established in the context of sub-level set filtrations. 
Our result extends this framework to node-labeling-based filtrations we designed in the context of graphs, addressing the challenges of incorporating target nodes in LP tasks. 
It guarantees that small changes in the node labeling function lead to bounded changes in the resulting persistence diagrams, measured by the bottleneck distance. This result validates the robustness of persistence diagrams derived from node labeling functions, ensuring their reliability for LP tasks.
In particular, this theorem applies when the node labeling functions are defined as DRNL $f^{(u,v)}_{drnl}$ and Degree DRNL $f^{(u,v)}_{degdrnl}$, as shown in the following corollary.

\begin{corollary}\label{cor_1}
    Let $G^{(u,v)} = (V,E)$ be a graph with target nodes $u,v$ in $V$. Then, the bottleneck distance between two persistence diagrams $dgm_p(f^{(u,v)}_{drnl})$ and $dgm_p(f^{(u,v)}_{degdrnl})$ is bounded by 1. 
\end{corollary}
\begin{proof}
    By Theorem~\ref{thm:stability}, we have 
    \begin{align*}
        & D_B(dgm_p(f^{(u,v)}_{drnl}),dgm_p(f^{(u,v)}_{degdrnl}) )  \\
        & \indent\leq \lVert f^{(u,v)}_{drnl}-f^{(u,v)}_{degdrnl} \rVert_\infty = \max_w (M - \deg(w)) / M \le 1. 
    \end{align*}
\end{proof}

The boundedness established in Corollary~\ref{cor_1} demonstrates that DRNL and Degree DRNL yield similar persistence diagrams. 
This finding confirms that degree information is included while maintaining the information extracted using the DRNL.
\begin{corollary}
     Suppose there exists a graph isomorphism with target nodes between two graphs $G_1^{(u_1,v_1)}$ and $G_2^{(u_2,v_2)}$. 
     Then, the persistence diagrams of $G_1^{(u_1,v_1)}$ and $G_2^{(u_2,v_2)}$, obtained from filtrations constructed using the same fixed node labeling function, are identical.
\end{corollary}

The result guarantees consistency in the representation of isomorphic graphs that preserve target nodes, ensuring they produce identical persistence diagrams under the same node labeling function.

\subsection{Persistence Image}
\label{subsec:persistenceimage}

We convert the persistence diagram into a PI~\cite{adams2017persistence}. 
For a given persistence diagram \( D \),
consider a linear transform \( L: \mathbb{R}^2 \rightarrow \mathbb{R}^2 \) defined by \( L(x, y) = (x, y-x) \). 
The image set of \( D \) under this transformation is denoted as \( L(D) \). 
For each point $(b,d')$ in $L(D)$, a weight function \( \phi_{(b,d')}: \mathbb{R}^2 \rightarrow \mathbb{R} \) is defined that assigns a weight to each point in the persistence diagram. 
A common choice for \( \phi_{(b,d')} \) is the Gaussian function centered at $(b,d')$.
The nonnegative function is defined as $h:\mathbb{R}^2 \rightarrow \mathbb{R}$, as $h(x,y)=1/\log(1+ \lvert y \rvert)$.
The function $h$ is zero along the horizontal $x$-axis, and is continuous and piecewise differentiable, satisfying the conditions presented in~\cite{adams2017persistence}.
The persistence surface $\rho_D:\mathbb{R}^2 \rightarrow \mathbb{R}$ is defined as
\begin{equation} 
\rho_D(z) = \sum_{(b,d') \in L(D)} h(b,d')\phi_{(b,d')}(z). 
\end{equation}

The continuous surface \( \rho_D \) is discretized into a finite-dimensional representation over a predefined grid. 
This grid consists of \( n \) cells, each corresponding to a specific region in the plane. 
The PI is defined as an array of values \( I(\rho_D)_p \) for each cell $p$. 
Each $I(\rho_D)_p$ in this array is computed by integrating the persistence surface \( \rho_D \) over the area of cell $p$:
\begin{equation}
    I(\rho_D)_p  = \iint_{p} \rho_D \, dy \, dx.
\end{equation}

\subsection{Predicting the Existence of the Target Link}
\label{subsec:mlp}
For the given target nodes $(u,v)$, we sample the $(k,l)$-angle hop subgraph \( \mathcal{N}^{(k,l)}_{u,v}\), denoted as \(\mathcal{N}^-\) (Section~\ref{subsec:anglehop}), assuming that the target link does not exist during this process. 
On this subgraph, we extract topological features by calculating PH and its vectorization (i.e., the PI, as described in Sections~\ref{subsec:persistenthomology} and~\ref{subsec:persistenceimage}). The vectorization is calculated for each dimension and concatenated. If $k \neq l$, for symmetry, we repeat the same process with the $(l,k)$-angle hop subgraph once and consider the average of the two vectors, denoting this vector as $x^-$. 
To observe the difference in topological features, we consider a subgraph \( \mathcal{N}+ \) obtained by connecting the target link to \(\mathcal{N}^-\).
For this graph, \(x^+\) denotes the vector obtained using this method.

To predict the existence of the target link with the vectors \(x^-\) and \(x^+\), we employ an MLP classifier $\Phi: \mathbb{R}^{2(d+1)n^2} \rightarrow \mathbb{R}$ where $n$ represents the resolution of the PI, and $d$ denotes the maximal dimension of PH. 
The model predicts the existence of a link between two target nodes with the following probability: 
\begin{equation}
    z_{uv} = \sigma(\Phi(x)), 
\end{equation}
where $x$ is the concatenation of $x^-$ and $x^+$, and $\sigma$ is the activation function. 
For the training dataset \(\mathcal{X} \subseteq V \times V\), comprising positive and negative links corresponding to the elements of \(E\) and \((V\times V)\setminus E\), respectively, we define the loss function as follows:
\begin{equation}
    \sum_{(u,v) \in \mathcal{X}} BCE(z_{uv}, y_{uv}),
\end{equation}
where \(BCE(\cdot,\cdot)\) represents the binary cross-entropy loss and \(y_{uv}\) denotes the label of the target link \((u,v)\), which is \(0\) for negative links or \(1\) for positive links.

\subsection{The Interpretability of the Feature Extraction}
\label{subsec:interpretability}

Our method enhances the interpretability of feature extraction compared to GNN-based methods. While GNNs can extract feature vectors and visualize them using dimension reduction techniques, understanding the specific reasons behind the values of these features remains challenging. This difficulty stems from the complex and uncertain nature of the training process, which involves various factors such as optimization methods, learning rates, batch sizes, loss functions, and data distributions. In contrast, our method employs a predefined feature extraction process that operates independently of any training phase, as illustrated in Figure~\ref{fig:Blackbox}. This approach not only eliminates the ambiguities associated with training but also allows us to precisely understand what each component of the extracted feature vector represents and identify the critical aspects of our method that influence these values. This level of clarity is particularly valuable for applications requiring transparent and accountable decision-making processes.

For example, the development of the Degree DRNL node labeling was enabled by the inherent interpretability of our method. Initially, we utilized the DRNL to construct filtrations but noticed that the Power dataset exhibited the lowest accuracy among our benchmark datasets. This observation led us to conduct an extensive analysis of the subgraphs and their corresponding feature vectors within the Power dataset. As depicted in Figure~\ref{drnl}, we identified instances where structurally distinct subgraphs produced identical feature vectors, highlighting a limitation when utilizing DRNL. Despite one subgraph being labeled as 1 and the other being labeled as 0 according to the presence and absence of a target link, our method resulted in identical feature vectors for both subgraphs. 
We easily identified the reasons behind these identical vectors, which was straightforward since feature extraction does not have a training process.
To address this, we incorporated degree information into the node labeling, as detailed in Equation~\ref{eq:degdrnl}.
We refined the labeling by adding a fractional value between 0 and 1 based on the degree. This enhancement complements the DRNL by assigning an order according to degree, thereby prioritizing simplexes during their simultaneous addition in the filtration process.
Furthermore, we established a stability theorem that guarantees minor perturbations in node labelings result in bounded variations in persistence diagrams, thus enhancing the robustness and reliability of our method.

Our methodology not only facilitates a deep understanding of the feature extraction process but also allows for targeted modifications based on specific needs. This adaptability ensures that our approach can be finely tuned to enhance performance or address specific characteristics of the data set.

\subsection{Multiangle PHLP}
\label{subsec:maphlp}
The MA-PHLP maximizes the advantages of PHLP by examining data from various angles through the extraction of subgraphs based on a hyperparameter, the maximum hop (max hop, denoted as $H$).
The types of angles are elements of all combinations of \(k\) and \(l\) within the set \(\{(k,l) \in \mathbb{Z}^2 | 0 \leq l \leq k \leq H, k > 0\}\).
If we define the prediction probability of a PHLP for each type of angle hop as $z_i$ for $i=1, 2, ..., N$, then MA-PHLP predicts the likelihood of the link existence with the following probability: 
\begin{equation}
    p = \sum_{i=1}^N \alpha_iz_i ,
\end{equation}
where $\alpha = (\alpha_1,...,\alpha_N) \in \mathbb{R}^N$ is a trainable parameter.
We apply the softmax function to the parameter \(\alpha\) to ensure that the sum of all elements equals $1$. 
Moreover, MA-PHLP is trained using the binary cross-entropy loss.

\subsection{Hybrid Method}
\label{subsec:hybrid}
The proposed approach easily integrates with existing subgraph methods. Subgraph methods treat the LP task as a binary classification problem comprising two components: a feature extractor $F$ and classifier $P$. Vectors with PH information calculated using the proposed methods are incorporated through concatenation before the classifier.
The detailed process of the hybrid method is outlined as follows: 
\begin{enumerate}
    \item \textbf{Subgraph Extraction:} For the given graph $G$ and target nodes $(u,v)$, $k$-hop subgraph $\mathcal{N}^k_{u,v}$ is extracted. 
    \item \textbf{Feature Extraction:} Existing methods extract features $Z = F(\mathcal{N}^k_{u,v})$ from the subgraph.
    \item \textbf{Persistent Image Calculation:} The methods described in Sections~\ref{subsec:filtration}, \ref{subsec:persistenthomology}, and~\ref{subsec:persistenceimage} are applied to $\mathcal{N}^k_{u,v}$, where $I$ denotes the PI vector. An MLP $\Phi:\mathbb{R}^m \rightarrow \mathbb{R}^n$ transforms the PI into a format similar to $Z$. 
    For the hybrid method of MA-PHLP, $\mathcal{N}^k_{u,v}$ is replaced with multiangle subgraphs, concatenating their PI vectors.
    \item \textbf{Classification:} Next, $\alpha_1 Z$ and $\alpha_2 \Phi(I)$ are concatenated, where $\alpha_1$ and $\alpha_2$ are trainable parameters. The softmax function is applied to the parameter \(\alpha=(\alpha_1,\alpha_2)\), ensuring that the sum of elements equals $1$, denoted by $J$.
    This concatenated vector is classified using the existing method's classifier, $P(J)$.
\end{enumerate}
\begin{table*}[!htbp]
\centering
\caption{Link prediction performance measured by the AUC on Benchmark datasets (90\% observed links)}
\resizebox{\linewidth}{!}{
\begin{tabular}{l|cccccccccc}
\toprule
Dataset & USAir & NS & PB & Yeast & C.~ele & Power & Router & E.~coli \\
\midrule
AA & $95.06 \pm 1.03$ & $94.45 \pm 0.93$ & $92.36 \pm 0.34$ & $89.43 \pm 0.62$ & $86.95 \pm 1.40$ & $58.79 \pm 0.88$ & $56.43 \pm 0.51$ & $95.36 \pm 0.34$ \\
Katz & $92.88 \pm 1.42$ & $94.85 \pm 1.10$ & $92.92 \pm 0.35$ & $92.24 \pm 0.61$ & $86.34 \pm 1.89$ & $65.39 \pm 1.59$ & $38.62 \pm 1.35$ & $93.50 \pm 0.44$ \\
PR & $94.67 \pm 1.08$ & $94.89 \pm 1.08$ & $93.54 \pm 0.41$ & $92.76 \pm 0.55$ & $90.32 \pm 1.49$ & $66.00 \pm 1.59$ & $38.76 \pm 1.39$ & $95.57 \pm 0.44$ \\
WLK & $96.63 \pm 0.73$ & $98.57 \pm 0.51$ & $93.83 \pm 0.59$ & $95.86 \pm 0.54$ & $89.72 \pm 1.67$ & $82.41 \pm 3.43$ & $87.42 \pm 2.08$ & $96.94 \pm 0.29$ \\
WLNM & $95.95 \pm 1.10$ & $98.61 \pm 0.49$ & $93.49 \pm 0.47$ & $95.62 \pm 0.52$ & $86.18 \pm 1.72$ & $84.76 \pm 0.98$ & $94.41 \pm 0.88$ & $97.21 \pm 0.27$ \\
\midrule
N2V & $91.44 \pm 1.78$ & $91.52 \pm 1.28$ & $85.79 \pm 0.78$ & $93.67 \pm 0.46$ & $84.11 \pm 1.27$ & $76.22 \pm 0.92$ & $65.46 \pm 0.86$ & $90.82 \pm 1.49$ \\
SPC & $74.22 \pm 3.11$ & $89.94 \pm 2.39$ & $83.96 \pm 0.86$ & $93.25 \pm 0.40$ & $51.90 \pm 2.57$ & $91.78 \pm 0.61$ & $68.79 \pm 2.42$ & $94.92 \pm 0.32$ \\
MF & $94.08 \pm 0.80$ & $74.55 \pm 4.34$ & $94.30 \pm 0.53$ & $90.28 \pm 0.69$ & $85.90 \pm 1.74$ & $50.63 \pm 1.10$ & $78.03 \pm 1.63$ & $93.76 \pm 0.56$ \\
LINE & $81.47 \pm 10.71$ & $80.63 \pm 1.90$ & $76.95 \pm 2.76$ & $87.45 \pm 3.33$ & $69.21 \pm 3.14$ & $55.63 \pm 1.47$ & $67.15 \pm 2.10$ & $82.38 \pm 2.19$ \\
\midrule
SEAL & $\textcolor{blue}{97.10 \pm 0.87}$ & $98.25 \pm 0.61$ & $\textcolor{violet}{95.07 \pm 0.39}$ & $97.60 \pm 0.33$ & $89.54 \pm 1.23$ & $86.21 \pm 2.89$ & ${95.07 \pm 1.63}$ & $97.57 \pm 0.30$ \\
WP & $\ \textcolor{red}{98.20\pm0.57}$ & $\ \textcolor{red}{99.12\pm0.45}$ & $\ \textcolor{red}{95.42\pm0.25}$ & $\ \textcolor{red}{98.21\pm0.17}$ & $\ \textcolor{red}{93.30\pm0.91}$ & $\textcolor{violet}{92.11 \pm 0.76}$ & $\ \textcolor{red}{97.15\pm0.29}$ & $\ \textcolor{red}{98.54\pm0.19}$ \\
LGLP   & $97.09 \pm 0.13$ & \textcolor{red}{$99.12 \pm 0.00$} & $94.70 \pm 0.04$ & $97.53 \pm 0.13$ & $88.64 \pm 0.29$ & $85.63 \pm 0.07$ & $95.51 \pm 0.07$ & $\textcolor{blue}{98.39 \pm 0.08}$ \\
MPLP   & $97.01 \pm 0.54$ & $96.17 \pm 0.84$ & $94.06 \pm 0.58$ & $94.25 \pm 0.43$ & $\textcolor{blue}{90.48 \pm 0.87}$ & $73.71 \pm 1.08$ & $91.90 \pm 0.50$ & $96.67 \pm 0.14$ \\
\midrule
MA-PHLP & $\textcolor{blue}{ 97.10 \pm 0.69 }$ & $\textcolor{violet}{98.88\pm0.45}$ & $\textcolor{blue}{95.10\pm0.26}$ & $\textcolor{blue}{97.98\pm0.22}$ & $\textcolor{violet}{90.33\pm1.16}$ & $\textcolor{blue}{93.05\pm0.45}$ & $\textcolor{blue}{96.30\pm0.43}$ & $97.64\pm0.20$  \\
MA-PHLP (dim$0$) & $97.10\pm0.73$ & $98.78\pm0.65$ & $95.06\pm0.28$ & $\textcolor{blue}{97.98\pm0.23}$ & $89.88\pm1.22$ & $\ \textcolor{red}{93.37} \pm\ \textcolor{red}{0.41}$ & $\textcolor{blue}{96.37\pm0.43}$ & $\textcolor{violet}{97.72\pm0.17}$  \\
\bottomrule
\end{tabular}}
\label{tbl:MA-PHLP}
\end{table*}

\section{Experiments}
This section evaluates the performance of MA-PHLP. 
The experiments were also conducted using only zero-dimensional homology (MA-PHLP (dim$0$)).
We used the area under the curve (AUC)~\cite{bradley1997use} as an evaluation metric. We repeated all experiments 10 times and reported the mean and standard deviation of the AUC values. 
The code for our implementation is available at \href{https://github.com/AI-hew-math/MA-PHLP}{https://github.com/AI-hew-math/MA-PHLP}.

\subsection{Experimental Settings}

\noindent\textbf{Baselines.} To evaluate the effectiveness of PHLP, we compared the proposed model with five heuristic methods, four embedding-based methods, and two GNN-based models. The heuristic methods include the Adamic-Adar (AA)~\cite{adamic2003friends}, Katz index (Katz)~\cite{katz1953new}, PageRank (PR)~\cite{brin1998anatomy}, Weisfeiler-Lehman graph kernel (WLK)~\cite{shervashidze2011weisfeiler}, and Weisfeiler-Lehman neural machine (WLNM)~\cite{zhang2017weisfeiler}. 
For the embedding-based methods, we applied N2V~\cite{grover2016node2vec}, spectral clustering (SPC)~\cite{tang2011leveraging}, matrix factorization (MF)~\cite{koren2009matrix}, and LINE~\cite{tang2015line}. Moreover, SEAL~\cite{zhang2018link}, WP~\cite{pan2021neural}, {LGLP~\cite{cai2022line} and MPLP~\cite{dong2024pure} represent the GNN-based methods. 

\noindent\textbf{Datasets.}
In line with previous studies~\cite{zhang2018link} and~\cite{pan2021neural}, we evaluate the performance of our MA-PHLP on the eight datasets in Table~\ref{tbl:dataset} without node attributes: USAir~\cite{batagelj2006pajek}, NS~\cite{newman2006finding}, PB~\cite{ackland2005mapping}, Yeast~\cite{von2002comparative}, C.~elegans (C.~ele)~\cite{watts1998collective}, Power~\cite{watts1998collective}, Router~\cite{spring2002measuring}, and E.~coli~\cite{zhang2018beyond}. 
The detailed statistics for each dataset are summarized in Table~\ref{tbl:dataset}.

\begin{table}[!htbp]
\centering
\caption{Statistics of the datasets}
\begin{tabular}{l|ccccc}
\toprule
Dataset & \#Nodes & \#Edges & Avg. node deg. & Density \\
\midrule
{USAir} & 332 & 2126 & 12.81 & 3.86e-2 \\ 
{NS} & 1589 & 2742 & 3.45 & 2.17e-3 \\ 
{PB} & 1222 & 16714 & 27.36 & 2.24e-2 \\ 
{Yeast} & 2375 & 11693 & 9.85 & 4.15e-3 \\ 
{C.ele} & 297 & 2148 & 14.46 & 4.87e-2 \\ 
{Power} & 4941 & 6594 & 2.67 & 5.40e-4 \\ 
{Router} & 5022 & 6258 & 2.49 & 4.96e-4 \\ 
{E.coli} & 1805 & 15660 & 16.24 & 9.61e-3 \\ 
\bottomrule
\end{tabular}
\label{tbl:dataset}
\end{table}

\noindent\textbf{Implementation Details.} 
All edges in the datasets were split into training, validation, and testing datasets with proportions of $0.85$, $0.05$, and $0.1$, respectively, ensuring a fair comparison with previous studies.
The max hop $M$ was set to $3$ for most datasets (Table~\ref{tbl:MA-PHLP}). 
However, for the E.~coli dataset, it was reduced to $2$ when employing one-dimensional homology due to memory constraints. 
Conversely, for the Power dataset, the max hop was set to $7$ because it does not demand heavy memory and computation time.
The sigmoid function was employed for the activation function of the PHLP classifier.
Tables~\ref{tbl:PHLP+SEAL} and~\ref{tbl:PHLP+WP} present the results of the hybrid methods using SEAL~\cite{zhang2018link} and WP~\cite{pan2021neural}, respectively. 
We choose them because SEAL is simple and powerful GNN method and WP show the hightest scores on 7 datasets.
For these experiments, a two-layer MLP was used for the MLP $\Phi$ in Step $3$ of Section~\ref{subsec:hybrid}. 
We set the $k$-hops following the original methods, SEAL and WP, and the max hops $M$ of MA-PHLP were set as the $k$, except for the Power dataset.
For the Power dataset, we set the $k$-hop to $1$-hop and max hop $M$ to $7$, respectively, which is discussed in detail in Section~\ref{hop_analysis}.

\subsection{Results}
\noindent\textbf{Results of MA-PHLP.}
Table~\ref{tbl:MA-PHLP} presents the AUC scores for each model on the benchmark datasets. 
The top three models are colored by} \textcolor{red}{First}, \textcolor{blue}{Second}, \textcolor{violet}{Third}.
The results of AA, Katz, WLK, WLNM, N2V, SPC, MF, LINE, and SEAL are copied from SEAL~\cite{zhang2018link} for comparison. 
The MA-PHLP demonstrates high performance across most datasets, achieving competitive scores.
The proposed model outperforms several baselines, falling between the GNN-based models in terms of the AUC score. 
Notably, for the Power dataset, MA-PHLP achieves the highest AUC score, indicating its effectiveness in capturing link patterns.

\noindent\textbf{Results of Hybrid Methods.}
\begin{table}[!htbp]
\centering
\caption{AUC scores for SEAL with and without TDA features}
\begin{tabular}{l|cccccccccc}
\toprule
Dataset & SEAL & MA-PHLP + SEAL \\
\midrule
USAir & $97.10 \pm 0.87$ & $\mathbf{97.41} \pm \mathbf{0.62}$ \\
NS & $98.25 \pm 0.61$ & $\mathbf{98.97} \pm \mathbf{0.30}$ \\
PB & $95.07\pm0.39$ & $\mathbf{95.14} \pm \mathbf{0.39}$ \\
Yeast & $97.60\pm0.33$ & $\mathbf{97.93} \pm \mathbf{0.18}$ \\
C.ele & $89.54 \pm 1.23$ & $\mathbf{89.61} \pm \mathbf{1.12}$ \\
Power & $86.21 \pm 2.89$ & $\mathbf{95.53} \pm \mathbf{0.33}$ \\
Router & $95.07 \pm 1.63$ & $\mathbf{96.15} \pm \mathbf{1.26}$ \\
E.coli & $97.57\pm0.30$ & $\mathbf{97.93}\pm\mathbf{0.34}$ \\
\bottomrule
\end{tabular}\label{tbl:PHLP+SEAL}
\end{table}
Simply concatenating the PI vector calculated using PHLP with the final output of the SEAL model increases AUC scores for all datasets, as listed in Table~\ref{tbl:PHLP+SEAL}. This outcome suggests that when the SEAL model lacks topological information for inference, the vectors calculated using PHLP can serve as additional inputs.

\begin{table}[!htbp]
\centering
\caption{AUC scores for WALKPOOL (WP) with and without TDA features}
\begin{tabular}{l|cccccccccc}
\toprule
Dataset & WP & MA-PHLP + WP \\
\midrule
USAir & $ 98.20\pm0.57 $ & $\mathbf{98.27} \pm \mathbf{0.53}$ \\
NS & $ 99.12\pm0.45 $ & $\mathbf{99.24} \pm \mathbf{0.32}$ \\
PB & $ 95.42\pm0.25 $ & $\mathbf{95.58} \pm \mathbf{0.32}$ \\
Yeast & $ 98.21\pm0.17 $ & $\mathbf{98.25} \pm \mathbf{0.18}$ \\
C.ele & $ 93.30\pm0.91 $ & $\mathbf{93.32} \pm \mathbf{0.71}$ \\ 
Power & $ 92.11\pm0.76 $ & $\mathbf{96.09} \pm \mathbf{0.38}$ \\
Router & $ 97.15\pm0.29 $ & $\mathbf{97.18} \pm \mathbf{0.24}$ \\
E.coli & $ 98.54\pm0.19 $ & $\mathbf{98.57} \pm \mathbf{0.20}$ \\
\bottomrule
\end{tabular}\label{tbl:PHLP+WP}
\end{table}
Similarly, we attempted to hybridize PHLP with the current SOTA model, WP. 
As presented in Table~\ref{tbl:PHLP+WP}, a slight increase in AUC scores is observed for all datasets. The Power dataset demonstrates significant improvement.
 
\subsection{Ablation Study}\label{subsec:Ablation Study}
\noindent\textbf{Effects of Degree DRNL.} To assess the proposed Degree DRNL regarding the influence of incorporating degree information on model performance, we conducted experiments using DRNL and Degree DRNL and compared the results. 
\begin{table}[!htbp]
\centering
\caption{AUC scores for MA-PHLP (dim$0$) by node labeling}
\begin{tabular}{l|cccccccccc}
\toprule
Dataset & DRNL & Degree DRNL \\
\midrule
USAir & $96.73\pm0.64$ & $\mathbf{97.10} \pm \mathbf{0.73}$ \\
NS & ${98.35} \pm {0.58}$ & $\mathbf{98.78}\pm\mathbf{0.65}$ \\ 
PB & $94.49\pm0.27$ & $\mathbf{95.06} \pm \mathbf{0.28}$ \\ 
Yeast & $97.42\pm0.27$ & $\mathbf{97.98} \pm \mathbf{0.23}$ \\
C.ele& $88.97\pm1.37$ & $\mathbf{89.88} \pm \mathbf{1.22}$ \\
Power & $88.51\pm0.81$ & $\mathbf{92.77} \pm \mathbf{0.47}$ \\
Router & $96.21\pm0.53$ & $\mathbf{96.37} \pm \mathbf{0.43}$ \\ 
E.coli & $97.15\pm0.18$ & $\mathbf{97.72} \pm \mathbf{0.17}$ \\
\bottomrule
\end{tabular}
\label{tbl:nodelabel}
\end{table}
We used MA-PHLP (dim$0$) for the experiments.
Table~\ref{tbl:nodelabel} presents the AUC scores of MA-PHLP (dim$0$) with DRNL and Degree DRNL.
Across all datasets, MA-PHLP (dim$0$) yields higher AUC scores when used with Degree DRNL than with DRNL.
The substantial improvement observed in the Power dataset is noteworthy, where Degree DRNL yields an increase of over $4$ points in the AUC score.
These experiments demonstrate the importance of incorporating degree information into node labeling, revealing its efficacy in enhancing the performance of MA-PHLP.

\begin{table}[!htbp]
\centering
\caption{AUC scores for PHLP (dim$0$) with various $(k,l)$-angle hops}
\begin{tabularx}{0.44\textwidth}{l|YY}
\toprule
Dataset & (1,0) & (1,1) \\
\midrule
USAir & $\mathbf{96.15}\pm\mathbf{0.83}$ & $95.87\pm0.83$ \\
NS & $98.28\pm0.55$ & $\mathbf{98.66}\pm\mathbf{0.66}$ \\
PB & $93.95\pm0.34$ & $\mathbf{94.46}\pm\mathbf{0.36}$ \\
Yeast & $95.52\pm0.32$ & $\mathbf{97.31}\pm\mathbf{0.20}$ \\
C.ele & $86.18\pm2.12$ & $\mathbf{87.57}\pm\mathbf{1.20}$ \\
Power & $73.39\pm0.99$ & $\mathbf{77.83}\pm\mathbf{1.44}$ \\
Router & $92.09\pm0.57$ & $\mathbf{93.25}\pm\mathbf{0.47}$ \\
E.coli & $96.94\pm0.24$ & $\mathbf{96.95}\pm\mathbf{0.28}$ 
\end{tabularx}

\begin{tabularx}{0.44\textwidth}{l|YYY}
\toprule
Dataset & (2,0) & (2,1) & (2,2) \\
\midrule
USAir & $96.69\pm0.92$ & $96.74\pm0.84$ & $\mathbf{96.85}\pm\mathbf{0.83}$ \\
NS & $\mathbf{98.72}\pm\mathbf{0.51}$ & $98.59\pm0.65$ & $98.56\pm0.47$ \\
PB & $94.78\pm0.30$ & $94.73\pm0.30$ & $\mathbf{94.82}\pm\mathbf{0.24}$ \\
Yeast & $\mathbf{97.71}\pm\mathbf{0.18}$ & $97.66\pm0.27$ & $97.58\pm0.28$ \\
C.ele & $88.86\pm1.48$ & $\mathbf{89.16}\pm\mathbf{1.31}$ & $89.08\pm1.07$ \\
Power & $80.27\pm1.07$ & $83.90\pm1.29$ & $\mathbf{86.12}\pm\mathbf{0.86}$ \\
Router & $95.65\pm0.44$ & $\mathbf{95.71}\pm\mathbf{0.39}$ & $94.51\pm0.69$ \\
E.coli & $97.26\pm0.16$ & $97.29\pm0.24$ & $\mathbf{97.41}\pm\mathbf{0.21}$ \\
\bottomrule
\end{tabularx}
\label{tbl:angle}
\end{table}

\noindent\textbf{Angles of PHLP.} Table~\ref{tbl:angle} presents the performance of PHLP (dim $0$) concerning various \((k,l)\)-angle hop subgraphs.
Section~\ref{subsec:anglehop} proposed angle hop subgraphs as an alternative to traditional $k$-hop subgraphs to capture information from various perspectives. 
Moreover, MA-PHLP is proposed to aggregate information from multiple angles. 
To investigate performance when extracting information from specific angles, we conducted experiments using PHLP at different angles.
We used only zero-dimensional PIs for the experiments.
Overall, the results demonstrate that the performance is favorable for cases corresponding to the $k$-hop subgraph (where $k$ and $l$ are the same). 
However, some datasets perform better when $k$ and $l$ differ, highlighting the importance of varying angles to achieve the best performance. Therefore, using MA-PHLP is recommended to maximize performance consistently across datasets.

\noindent\textbf{Comparison with TLC-GNN.}
To demonstrate that the proposed method extracts superior topological information compared to the conventional TLC-GNN approach, we conducted the same experiments. 
The TLC-GNN was constructed by augmenting the graph convolutional network (GCN) model with PI information. 
We replaced the PI component of the TLC-GNN model with the PI vector produced by MA-PHLP, resulting in the MA-PHLP-GNN. 
The zero-dimensional PH was employed in this study for fair comparison because TLC-GNN used only zero-dimensional PH.
Additionally, we conducted experiments where the PH vectors were replaced with zero vectors, denoted as GCN. 
Table~\ref{tbl:tlcgnn} presents the experimental results.
\begin{table}[!htbp]
\centering
\caption{comparison of AUC scores with TLC-GNN}
\resizebox{0.95\linewidth}{!}{
\begin{tabular}{l|cccccccccc}
\toprule
Dataset & GCN & TLC-GNN & MA-PHLP-GNN \\ 
\midrule
Cora & $92.20\pm0.83$ & $\mathbf{93.16}\pm\mathbf{0.56}$ & $93.14\pm0.93$ \\ 
CiteSeer & $86.52\pm1.29$ & $87.38\pm0.97$ & $\mathbf{92.08}\pm\mathbf{0.53}$ \\ 
PubMed & $96.63\pm0.15$ & $96.30\pm0.25$ & $\mathbf{98.07}\pm\mathbf{0.07}$ \\ 
\bottomrule
\end{tabular}}
\label{tbl:tlcgnn}
\end{table}

The TLC-GNN is employed when the given data includes node attributes.
Hence, we conducted experiments using the following widely used benchmark datasets with node attributes: Cora~\cite{mccallum2000automating}, CiteSeer~\cite{giles1998citeseer}, and PubMed~\cite{namata2012query}. 
The MA-PHLP-GNN outperformed the TLC-GNN significantly on the CiteSeer and PubMed datasets while achieving similar performance on the Cora dataset. 
The TLC-GNN does not exhibit performance improvement for the PubMed dataset despite adding topological information. 
However, the proposed MA-PHLP-GNN demonstrates substantial performance enhancement.
Although the proposed model is developed for datasets without node attributes, it exhibits effective performance on datasets with node attributes through hybridization with the existing methods: SEAL+PHLP, WP+PHLP, and MA-PHLP-GNN. 
These experiments verify the versatility and effectiveness of this approach across diverse datasets.

\subsection{The hops and max hops of the hybrid methods}
\label{hop_analysis}

\begin{table*}[!htbp]
\centering
\caption{AUC scores on the power dataset varying $k$-hop and max hop $M$ of the hybrid methods}
\label{tab:hybrid_grid}
\renewcommand{\arraystretch}{1.2} 
\resizebox{0.9\linewidth}{!}{
\begin{tabular}{c|ccccccccc}
\toprule
\multicolumn{2}{c|}{} & \multicolumn{7}{c}{MA-PHLP (with max hop $M$)} \\
\midrule
\multicolumn{2}{c|}{$M$}& $1$ & $2$ & $3$ & $4$ & $5$ & $6$ & $7$\\
\midrule
\multirow{8}{*}{\rotatebox{90}{SEAL (with $k$-hop)}}& \multicolumn{1}{c|}{$k$}&\multicolumn{3}{c}{not robust to $k$} & \multicolumn{4}{|c}{robust to $k$} \\
& \multicolumn{1}{c|}{$1$} & $86.66 \pm 0.56$ & $90.22 \pm 0.79$ & $92.63 \pm 0.54$ & \multicolumn{1}{|c}{$94.50 \pm 0.41$} & $95.12 \pm 0.40$ & $95.46 \pm 0.38$ & $\textbf{95.53 $\pm$ 0.33}$ \\
& \multicolumn{1}{c|}{$2$} & $91.40 \pm 0.88$ & $90.20 \pm 0.80$ & $92.50 \pm 0.59$ & \multicolumn{1}{|c}{$94.39 \pm 0.39$} & $95.00 \pm 0.46$ & $95.31 \pm 0.40$ & $\textbf{95.39 $\pm$ 0.36}$ \\
& \multicolumn{1}{c|}{$3$} & $93.21 \pm 0.64$ & $92.79 \pm 0.60$ & $92.57 \pm 0.58$ & \multicolumn{1}{|c}{$94.22 \pm 0.43$} & $94.86 \pm 0.42$ & $\textbf{95.21 $\pm$ 0.45}$ & $95.19 \pm 0.44$ \\
& \multicolumn{1}{c|}{$4$} & $94.51 \pm 0.58$ & $94.23 \pm 0.34$ & $94.21 \pm 0.41$ & \multicolumn{1}{|c}{$94.31 \pm 0.40$} & $94.80 \pm 0.37$ & $95.10 \pm 0.33$ & $\textbf{95.27 $\pm$ 0.36}$ \\
& \multicolumn{1}{c|}{$5$} & $94.73 \pm 0.56$ & $94.45 \pm 0.44$ & $94.61 \pm 0.51$ & \multicolumn{1}{|c}{$94.80 \pm 0.53$} & $94.91 \pm 0.54$ & $95.13 \pm 0.51$ & $\textbf{95.19 $\pm$ 0.46}$ \\
& \multicolumn{1}{c|}{$6$} & $94.58 \pm 0.94$ & $94.81 \pm 0.32$ & $94.87 \pm 0.42$ & \multicolumn{1}{|c}{$95.06 \pm 0.50$} & $95.11 \pm 0.46$ & $\textbf{95.25 $\pm$ 0.45}$ & $95.25 \pm 0.46$ \\
& \multicolumn{1}{c|}{$7$} & $93.97 \pm 0.73$ & $94.22 \pm 0.35$ & $94.43 \pm 0.44$ & \multicolumn{1}{|c}{$94.78 \pm 0.45$} & $94.92 \pm 0.39$ & $\textbf{94.99 $\pm$ 0.52}$ & $94.98 \pm 0.39$ \\
\midrule
\multirow{8}{*}{\rotatebox{90}{WP (with $k$-hop)}}& \multicolumn{1}{c|}{$k$}& \multicolumn{2}{c}{not 
robust to $k$} & \multicolumn{5}{|c}{robust to $k$} \\
& \multicolumn{1}{c|}{$1$} & $87.53 \pm 0.73$ & $91.48 \pm 0.64$ & \multicolumn{1}{|c}{$93.55 \pm 0.48$} & $94.84 \pm 0.43$ & $95.53 \pm 0.46$ & $95.88 \pm 0.31$ & $\textbf{96.09 $\pm$ 0.38}$ \\
& \multicolumn{1}{c|}{$2$} & $92.51 \pm 0.58$ & $91.59 \pm 0.77$ & \multicolumn{1}{|c}{$93.49 \pm 0.58$} & $94.83 \pm 0.53$ & $95.56 \pm 0.59$ & $95.88 \pm 0.38$ & $\textbf{96.06 $\pm$ 0.45}$ \\
& \multicolumn{1}{c|}{$3$} & $94.04 \pm 0.46$ & $93.07 \pm 0.67$ & \multicolumn{1}{|c}{$93.61 \pm 0.52$} & $94.86 \pm 0.54$ & $95.61 \pm 0.60$ & $95.86 \pm 0.40$ & $\textbf{96.00 $\pm$ 0.52}$ \\
& \multicolumn{1}{c|}{$4$} & $93.55 \pm 0.71$ & $92.61 \pm 0.76$ & \multicolumn{1}{|c}{$93.68 \pm 0.55$} & $94.85 \pm 0.55$ & $95.59 \pm 0.58$ & $95.87 \pm 0.38$ & $\textbf{96.03 $\pm$ 0.45}$ \\
& \multicolumn{1}{c|}{$5$} & $93.40 \pm 0.70$ & $92.64 \pm 0.69$ & \multicolumn{1}{|c}{$93.66 \pm 0.53$} & $94.84 \pm 0.54$ & $95.55 \pm 0.59$ & $95.85 \pm 0.39$ & $\textbf{96.04 $\pm$ 0.52}$ \\
& \multicolumn{1}{c|}{$6$} & $93.34 \pm 0.75$ & $92.66 \pm 0.72$ & \multicolumn{1}{|c}{$93.64 \pm 0.55$} & $94.91 \pm 0.57$ & $95.55 \pm 0.58$ & $95.85 \pm 0.44$ & $\textbf{95.98 $\pm$ 0.55}$ \\
& \multicolumn{1}{c|}{$7$} & $93.30 \pm 0.73$ & $92.61 \pm 0.69$ & \multicolumn{1}{|c}{$93.65 \pm 0.56$} & $94.87 \pm 0.56$ & $95.56 \pm 0.58$ & $95.90 \pm 0.39$ & $\textbf{96.01 $\pm$ 0.52}$ \\
\bottomrule   
\end{tabular}}
\end{table*}

Determining the hyperparameters such as ``hop" and ``max hop" is crucial for the performance of the hybrid method. We conducted experiments to explore the effects of different combinations of these parameters. 
Given that the hybrid methods (e.g., MA-PHLP + SEAL and MA-PHLP + WP) exhibited the highest performance improvement on the Power dataset, 
we conducted experiments on the Power dataset.
Table~\ref{tab:hybrid_grid} presents the AUC scores for varying hop (SEAL or WP) and max hop (MA-PHLP).
For each target node, while the SEAL and WP extract a $k$-hop subgraph, the MA-PHLP calculates the PIs based on a subgraph with max hop $M$.
When the parameter \( M \) is \( 1 \) or \( 2 \), the AUC scores are not robust to \( k \), showing large variations; however, when \( M \) is \( 3 \), although MA-PHLP + SEAL still exhibits variations up to \(2\), MA-PHLP + WP shows only minor variations.
As $M$ exceeds $3$, the AUC scores of MA-PHLP + SEAL and MA-PHLP + WP are robust to $k$, exhibiting little sensitivity (maximum $0.84$) to variations.
This suggests that setting both the hop and the max hop to identical values may be permissible without further searching for optimal hyperparameters.
\begin{figure*}[!htbp]
\centering
   \includegraphics[width=0.9\linewidth]{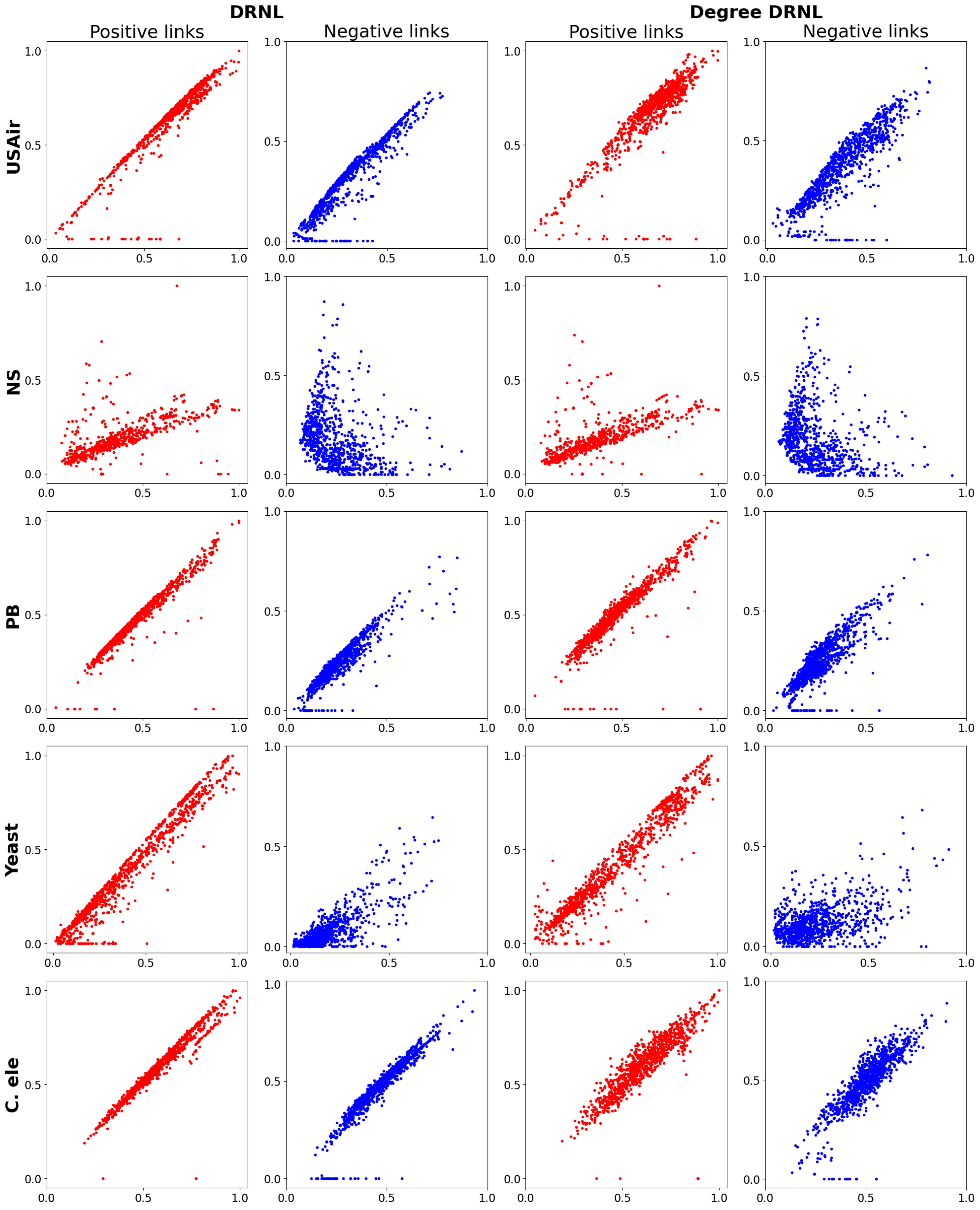}
   \hfil
   \caption{Visualization of vectors calculated using MA-PHLP (dim0). For each dataset, the first and second columns depict the projections of persistence images (PIs) when double radius node labeling (DRNL) is applied for node labeling, and the third and fourth columns represent the values obtained when Degree DRNL is applied. 
   The first and third columns plot the values produced from positive edges (i.e., target nodes labeled $1$), and the second and fourth columns plot the values produced from negative edges (i.e., target nodes labeled $0$). }
   \label{fig:posneg1}
\end{figure*}
\begin{figure*}[!htbp]
\centering
   \includegraphics[width=0.9\linewidth]{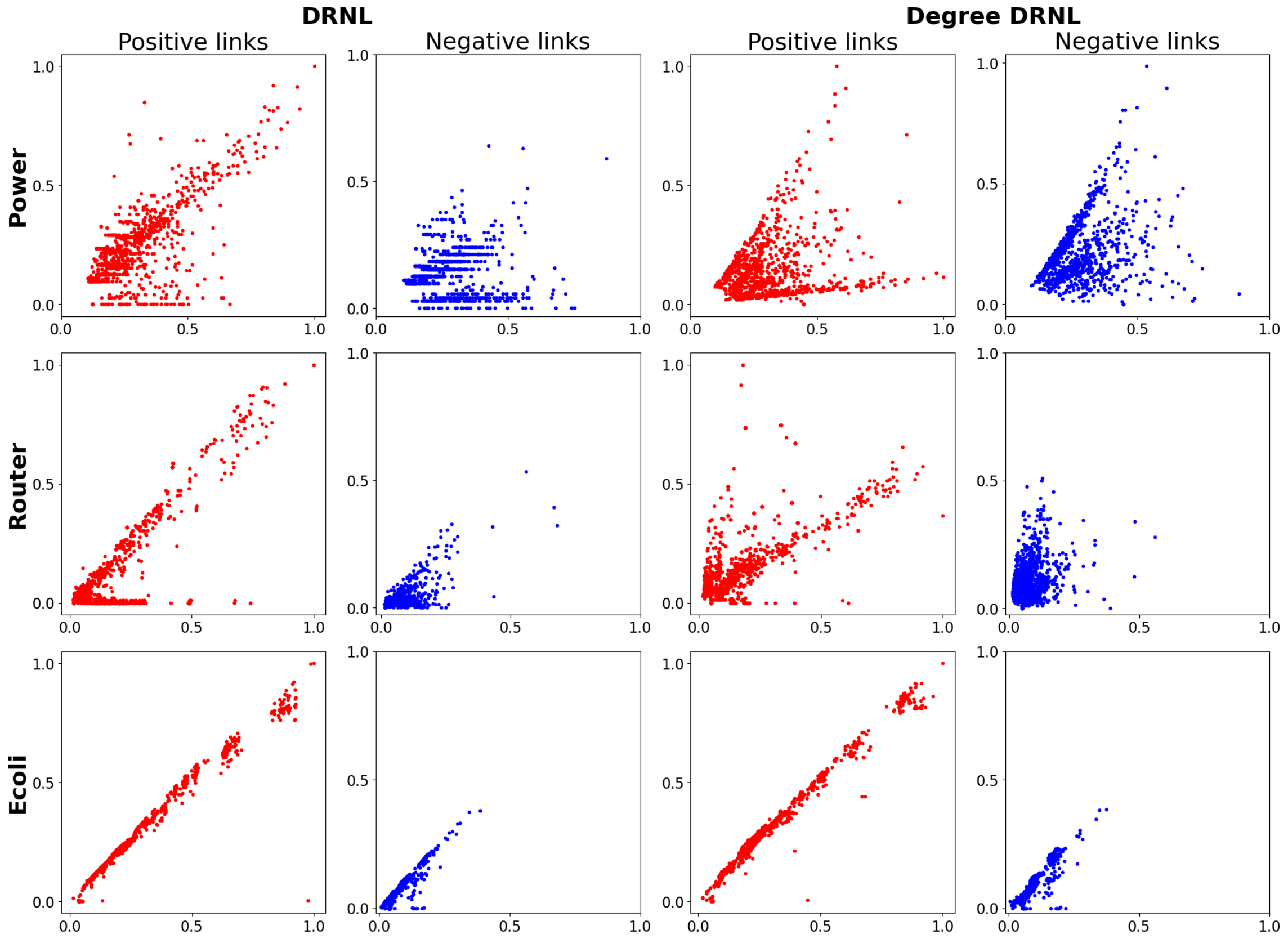}
   \hfil
   \caption{Visualization of vectors calculated using MA-PHLP (dim0).}
   \label{fig:posneg2}
\end{figure*}

\section{Analysis}

\subsection{Analysis of the PHLP}
Figs.~\ref{fig:posneg1} and~\ref{fig:posneg2} visualize concatenated PIs to illustrate how MA-PHLP (dim0) extracts topological features for LP. 
We let $\mathcal{Z} \subseteq \mathbb{R}^{2 \times k \times r^2}$ be a set of vectors calculated by MA-PHLP, where $k$ is the number of angles, and $r$ denotes the PI resolution. 
For $(z_1, z_2) \in \mathcal{Z}$, $z_1 \in \mathbb{R}^{k \times r^2}$ is the concatenation of PIs for all angles with a target link, and $z_2 \in \mathbb{R}^{k \times r^2}$ is the concatenation for cases without a target link.
We consider a function $h:\mathbb{R}^{k \times r^2} \rightarrow \mathbb{R}$ defined as 
$h(\vec{v}_1, ..., \vec{v}_k) = \frac{1}{k}\sum_{i=1}^k \lVert \vec{v}_i \rVert_1$, where $\vec{v}_i \in \mathbb{R}^{r^2}$ are PIs, and $\lVert \cdot \rVert_1$ denotes the $L_1$-norm.
For visualization, we transform $\mathcal{Z}$ into points in $\mathbb{R}^2$ using the function $G$, defined as $G(z_1,z_2) = (h(z_1),h(z_2))$ for each $(z_1, z_2) \in \mathcal{Z}$. 

We plot distributions of points separately for positive and negative links, considering both DRNL and Degree DRNL.
The distributions of the NS and Yeast datasets between positive and negative links display significant differences, supporting the highest performance in Table~\ref{tbl:nodelabel}.
In contrast, the distributions for the C.~ele and Power datasets are the most similar when using Degree DRNL, correlating with the lowest scores in Table~\ref{tbl:nodelabel}.

\subsection{Analysis of the Power Dataset}
In most LP models, including the SOTA models SEAL and WP, the Power dataset tends to have the lowest AUC scores among the datasets. 
In Table~\ref{tbl:MA-PHLP}, the Power dataset is at the bottom in terms of scores across models (e.g., WLK, WLNM, MF, LINE, SEAL, and WP).
However, the proposed model achieves the highest AUC scores on the Power dataset among baseline models, prompting an analysis of the reasons for this performance.

In Figure~\ref{fig:posneg2}, for DRNL, the Power dataset exhibits horizontal lines, indicating that the values $h(z_2)$ have a limited range of outcomes for vectors $z_2$ in cases without the target link; thus, the set of values $h(z_2)$ with the same value should be spread out.
This observation implies that, for numerous subgraphs the calculation of PIs yields similar outcomes despite the differences in their topological structures, posing a challenge in distinguishing between them.

To address this problem, we applied Degree DRNL, which incorporates degree information. 
The points in Figure~\ref{fig:posneg2} are distributed without horizontal lines, leading to the highest score increase, as listed in Table~\ref{tbl:nodelabel}.

The performance of heuristic methods, such as AA, Katz, and PR, tend to be similar to random guessing on datasets with low density, particularly in the cases of the Power and Router datasets. 
Embedding methods also display low performance. 
In contrast, the GNN-based methods demonstrate improved performance using subgraphs and the network learning ability.
However, the performance for the Power dataset is significantly lower than that for the Router dataset. 

\begin{table}[!htbp]
\centering
\caption{Average number of nodes in subgraphs for the Power and Router datasets}
\label{tab:num_nodes}
\begin{tabular}{l|cc|cc}
\toprule
& \multicolumn{2}{c|}{Power} & \multicolumn{2}{c}{Router} \\
\midrule
& positive & negative & positive & negative \\
\midrule
1-hop & 8.03 & 9.12 & 5.11 & 6.72 \\
2-hop & 22.26 & 24.85 & 29.21 & 13.94 \\
3-hop & 43.11 & 49.50 & 120.35 & 55.22 \\
4-hop & 71.72 & 82.16 & 411.87 & 176.34 \\
5-hop & 99.28 & 116.75 & 740.80 & 411.35 \\
6-hop & 136.23 & 158.27 & 1272.42 & 852.13 \\
7-hop & 182.22 & 210.35 & 1835.46 & 1498.58 \\
\bottomrule
\end{tabular}
\end{table}

To bridge this gap, we analyzed subgraphs with node labeling. 
The number of nodes within the selected subgraphs between positive and negative links was significantly different on the Router dataset but not the Power dataset (Table~\ref{tab:num_nodes}). 
This difference is attributed to the presence of the hub nodes in the Router dataset, which are connected to numerous nodes. 
Thus, the subgraphs corresponding to positive links tend to have more nodes than those corresponding to negative links. 

\begin{table}[!htbp]
\centering
\caption{Comparison of models by Max hop settings on the Power and Router datasets}
\label{tab:center}
\resizebox{\linewidth}{!}{
\begin{tabular}{l|l|cccc}
\toprule
\multirow{2}{*}{\rotatebox{90}{}}& Model & MA-PHLP & MA-PHLP & WP & MA-PHLP + WP \\
\cmidrule{2-6}
& Center & target & random & - & random \\
\midrule
\multirow{7}{*}{\rotatebox{90}{Power}} & 1-hop & $78.05\pm1.20$ & $85.66\pm0.86$ & $80.24\pm0.95$ & $87.53\pm0.73$ \\
& 2-hop & $86.34\pm1.04$ & $90.52\pm0.73$ & $89.40\pm1.00$ & $91.59\pm0.77$ \\
& 3-hop & $89.65\pm0.64$ & $91.90\pm0.58$ & $\mathbf{92.11}\pm\mathbf{0.77}$ & $93.61\pm0.52$ \\
& 4-hop & $91.38\pm0.53$ & $92.67\pm0.55$ & $91.67\pm0.80$ & $94.85\pm0.55$ \\
& 5-hop & $92.27\pm0.40$ & $93.06\pm0.44$ & $91.39\pm0.78$ & $95.55\pm0.59$ \\
& 6-hop & $92.77\pm0.47$ & $93.16\pm0.49$ & $91.55\pm0.83$ & $95.85\pm0.44$ \\
& 7-hop &$\mathbf{93.06} \pm\mathbf{0.43}$  & $\mathbf{93.37} \pm\mathbf{0.41}$  & $91.50 \pm 0.89$ & $\mathbf{96.01} \pm\mathbf{0.52}$\\
\midrule
\multirow{7}{*}{\rotatebox{90}{Router}} & 1-hop & $93.12 \pm 0.45$ & $93.40 \pm 0.46$ & $94.48 \pm 0.36$ & $94.83 \pm 0.41$ \\
& 2-hop & $95.96 \pm 0.40$ & $95.70 \pm 0.45$ & $97.15 \pm 0.27$ & $97.22 \pm 0.23$ \\
& 3-hop & $96.38 \pm 0.41$ & $96.11 \pm 0.43$ & $\mathbf{97.28} \pm \mathbf{0.24}$ & $\mathbf{97.42} \pm \mathbf{0.27}$ \\
& 4-hop & $96.45 \pm 0.40$ & $96.22 \pm 0.43$ & OOM\footnotemark & OOM \\
& 5-hop & $\mathbf{96.46} \pm \mathbf{0.42}$ & $\mathbf{96.24} \pm \mathbf{0.48}$ & OOM & OOM \\
& 6-hop & $96.44 \pm 0.45$ & $96.23 \pm 0.47$ & OOM & OOM \\
& 7-hop & $96.43 \pm 0.45$ & $96.19 \pm 0.49$  & OOM & OOM \\
\bottomrule
\end{tabular}}
\end{table}
However, the Power dataset does not have hub nodes, and the number of nodes in the subgraph of positive links remains small. 
\footnotetext{OOM denotes ``out of GPU memory''.}
We randomly changed the center nodes $(a,b)$ for node labeling $f^{(a,b)}_{\text{degdrnl}}(w)$ increasing the performance, as listed in Table~\ref{tab:center}. 
This outcome highlights that setting target nodes as the center nodes may not effectively analyze the topological structure in the case of small graphs. 
Furthermore, the performance for the Power dataset continues to increase with increasing hops (Table~\ref{tab:center}), achieving an AUC score of $95.87$, which is significantly better than that of $92.11$ for WP. 
\newpage
\section{Conclusion}
This paper proposes PHLP, an explainable method that applies PH to analyze the topological structure of graphs to overcome the limitations of GNN-based methods for LP.
By employing the proposed methods, such as angle hop subgraphs and Degree DRNL, PHLP improves the analysis of the topological structure of graphs.
The experimental results demonstrate that the proposed PHLP method achieves competitive performance across benchmark datasets, even SOTA performance, especially on the Power dataset. 
Additionally, when integrated with existing GNN-based methods, PHLP improves performance across all datasets.
By analyzing the topological information of the given graphs, PHLP addresses the limitations of GNN-based methods and enhances overall performance. 
As demonstrated, PHLP provides explainable algorithms without relying on complex deep learning techniques, providing insight into the factors that significantly influence performance for the LP problem of graph data. 

\noindent\textbf{Limitations and Future works}
Although hybrid methods consistently improve performance, they often sacrifice interpretability. Consequently, we utilize MA-PHLP to preserve interpretability. However, MA-PHLP achieves the highest scores only on the Power dataset. To address this limitation, future efforts will be dedicated to designing hybrid approaches that maintain interpretability. These developments will ensure that our enhancements not only enhance predictive accuracy but also preserve the clarity necessary for informed decision-making in LP tasks.

\section*{Acknowledgements}
This work was supported by the NRF under the grant number 2021R1A2C3009648 and partially by the NRF under the grant number MSIT (RS-2023-00219980).

\bibliography{reference}
\bibliographystyle{IEEEtran}

\appendices
\section{}\label{sec:proof}

\noindent\textbf{Proof of Theorem 1.} Let $G^{(u,v)}=(V,E)$ be a graph with target nodes $u,v$ in $V$ and let $f^{(u,v)}$ and $g^{(u,v)}$ be two node labeling functions based on $(u,v)$ defined on $G^{(u,v)}$.
For simplicity, we denote these functions as \(f\) and \(g\), respectively. 
Denote the edge-weight functions derived from $f$ and $g$ as $W(f)$ and $W(g)$, respectively.
The power set $\mathbb{X}$ of the vertex set $V$ is an abstract simplicial complex.
Consider the Rips filtration function \( h_f: \mathbb{X} \rightarrow \mathbb{R} \) defined as 
\[
    h_f(\sigma) = \max \{ W(f)(w,z) \mid \text{ edge } \{w,z\} \subseteq \sigma \}
\]
whenever \( p \geq 1 \) for \( p \)-simplex \( \sigma \in \mathbb{X} \), and defined as \( h_f(z) = 0 \) for any \( 0 \)-simplex \( z \).
Denote $h_f^{-1}((-\infty,\epsilon])$ as $K_{\epsilon}$ for $\epsilon \in \mathbb{R}$.
Then, the Rips filtration is obtained as  
\(
    K_{\epsilon_1} \hookrightarrow K_{\epsilon_2} \hookrightarrow \cdots \hookrightarrow K_{\epsilon_m} = \mathbb{X}
 \)
for $\epsilon_1 \le \epsilon_2 \le \cdots \le \epsilon_m $.
By the stability theorem of the persistence diagram~\cite{cohen2005stability}, we know that $D_B(dgm_p(f),dgm_p(g)) \leq \lVert {h_f} - {h_g}\rVert_\infty$.
    
Now, we have $\lVert h_{f} - h_{g} \rVert_{\infty} = \lvert (h_{f} - h_{g})(\sigma_*) \rvert$ for a $p$-simplex $\sigma_* \in \mathbb{X}$.
By the maximality of $h_{f}$ and $h_{g}$, there are edges $(w_f,z_f)$ and $(w_g,z_g) \subseteq \sigma_*$ such that $h_{f}(\sigma_*) = W(f)(w_f,z_f)$ and $h_{g}(\sigma_*) = W(g)(w_g,z_g)$.
Consider an interpolation map $\mathcal{I}(t) = (1-t)W(f) + tW(g)$ for a real number $t$, where $\mathcal{I}(0)=W(f)$ and $\mathcal{I}(1)=W(g)$. 
Then, define a function $\mathcal{H}(t) = \mathcal{I}(t)(w_f,z_f) - \mathcal{I}(t)(w_g,z_g)$ for real number $t$.
By the maximality of the functions $h_f$ and $h_g$, we have $\mathcal{H}(0) \ge 0$ and $\mathcal{H}(1) \le 0$. 
Given that \( \mathcal{H} \) is continuous on the closed interval \([0,1]\), the intermediate value theorem guarantees the existence of a real number \( t_0 \) within \([0,1]\) such that \( \mathcal{H}(t_0) = 0 \).
That is, if we write \( \mathcal{I}(t_0) \) as \( W_* \), we have \( W_*(w_f,z_f) = W_*(w_g,z_g) \).
Now, observe that 
\begin{align*}
    \lVert h_{f} - h_{g} \rVert_{\infty} &= \lvert (h_{f} - h_{g})(\sigma_*) \rvert \\
    &= \lvert W(f)(w_f,z_f)-W(g)(w_g,z_g) \rvert \\
    &\le \lvert W(f)(w_f,z_f)-W_*(w_f,z_f) \rvert \\
    & \indent + \lvert W_*(w_g,z_g)-W(g)(w_g,z_g) \rvert\\
    &\le \lVert W(f)-W_* \rVert_\infty + \lVert W_*-W(g) \rVert_\infty\\
    &= \lVert W(f)-W(g) \rVert_\infty
\end{align*}
The last equality is derived from the expression \( W_* = (1-t_0)W(f) + t_0W(g) \).
    
By the definition of infinity norm, there exist two vertices $w$ and $z$ such that $\lVert W(f)-W(g) \rVert_\infty = \lvert W(f)(w,z)-W(g)(w,z)\rvert$.
Without loss of generality, we can assume $f(w) > f(z)$.
Suppose $g(w) > g(z)$. Then we have $\lVert W(f)-W(g) \rVert_\infty = \lvert W(f)(w,z)-W(g)(w,z)\rvert = \lvert f(w)-g(w) \rvert \le \lVert f-g \rVert_\infty$.
    
Next, suppose $g(w) \le g(z)$. 
Then we have $\lVert W(f)-W(g) \rVert_\infty = \lvert f(w)-g(z) \rvert$.
If $f(w) - g(z) >0$, then $\lvert f(w)-g(z) \rvert = f(w) - g(z) \le f(w) - g(w) \le \lVert f-g \rVert_\infty$.
If $f(w) - g(z) \le 0$, then $\lvert f(w)-g(z) \rvert = g(z) - f(w) \le g(z) - f(z) \le \lVert f-g \rVert_\infty$.
Thus we have $\lVert W(f)-W(g) \rVert_\infty \le \lVert f-g \rVert_\infty$ and conclude $D_B(dgm_p(f),dgm_p(g)) \leq \lVert f-g \rVert_\infty$. \qed

\vspace{-1mm}
\section{}\label{sec:algorithm}
\noindent \textbf{Algorithm of PHLP.} We present the training algorithms for the PHLP and MA-PHLP, as shown in Algorithms~\ref{alg:phlp} and ~\ref{alg:maphlp}.
\begin{algorithm}[!htbp]
\renewcommand{\arraystretch}{1.3}
\caption{PHLP Training Algorithm}
\label{alg:phlp}
\footnotesize
    \begin{algorithmic}[1]
    \raggedright
        \State{\bfseries Input: } 
            Observed graph $G=(V,E)$;\\
            Classifier $\Phi$ with He-initialized weights $\theta$;\\
            Angle hop parameters $(k,l)$, maximum epochs $Epochs$, batch size $m$;
        \State {\bfseries Output: } 
            Trained classifier $\Phi$ with optimized weights $\theta$;
        \State Sample true target links $\{x_1,x_2, \cdots, x_n\}$ from observed links $E$ and label them as 1;
        \State Sample false target links $\{x_{n+1}, x_{n+2}, \cdots, x_{2n}\}$ from $V \times V \setminus E$ and label them as 0;
        \For{$i=1\cdots2n$}
        \State Extract $(k,l)$-angle hop subgraph $\mathcal{N}_{x_i}^{(k,l)}$ for target link $x_i$;
        \State Compute Degree DRNL and assign edge weights to graphs with and without the target link, $\mathcal{N}^+$ and $\mathcal{N}^-$;
        \State Compute persistence diagrams $D^+$ and $D^-$ for $\mathcal{N}^+$ and $\mathcal{N}^-$ using edge weights as filtration values;
        \State Vectorize $D^+$ and $D^-$ into persistence images $PI^+$ and $PI^-$;
        \State If $k \neq l$, repeat the process for $(l, k)$-angle hop subgraph and average the two vectors;
        \State Concatenate $PI^+$ and $PI^-$ to form the final vector $PI_i$;
        \EndFor
        \State Form training dataset $\mathcal{X} = \{ PI_1, PI_2, \cdots, PI_{2n}\}$ and corresponding labels $Y=\{1,\cdots,1,0,\cdots,0\}$;
        \For{$e=1\cdots Epochs$}
        \State Sample a random batch $\{pi_1, pi_2, \cdots, pi_m \}$ from $\mathcal{X}$ and corresponding labels $\{y_1, y_2, \cdots, y_m\}$;
        \State Predict labels from classifier $\Phi$: $p_1,p_2, \cdots, p_m$;
        \State Train the classifier using binary cross-entropy loss $BCE$ and update weights $\theta$:
        \begin{center}
        \normalsize
        {$\nabla_{\theta}( \sum\limits_{pi_i \in \mathcal{X}}BCE(p_i, y_i))$};
        \end{center}
        \EndFor
    \end{algorithmic}
\end{algorithm}

\begin{algorithm}[H]
\renewcommand{\arraystretch}{1.3}
\caption{MA-PHLP Training Algorithm}
\label{alg:maphlp}
\footnotesize
    \begin{algorithmic}[1]
    \raggedright
        \State{\bfseries Input: } 
            Observed graph $G=(V,E)$;\\
            Max hop $H$, maximum epochs $Epochs$, batch size $m$;\\
            Classifiers $\Phi = (\Phi_1,\cdots,\Phi_N)$ with He-initialized weights $\theta = (\theta_1,\cdots,\theta_N)$;\\
            Trainable parameter $\alpha=(\alpha_1,\cdots,\alpha_N)$, where $N=|\{(k,l) \in \mathbb{Z}^2 : 0 \leq l \leq k \leq H, k > 0\}|$;
        \State {\bfseries Output: } 
            Trained classifiers $\Phi$ with optimized weights $\theta$;\\
            Optimized parameter $\alpha$;
        \State Sample true target links $\{x_1,x_2, \cdots, x_n\}$ from observed links $E$ and label them as 1;
        \State Sample false target links $\{x_{n+1}, x_{n+2}, \cdots, x_{2n}\}$ from $V \times V \setminus E$ and label them as 0;
        \For{$i=1\cdots2n$}
        \For {$k=1\cdots H$}
        \For {$l=k\cdots H$}
        \State Extract $(k,l)$-angle hop subgraph $\mathcal{N}_{x_i}^{(k,l)}$ for target link $x_i$;
        \State Compute Degree DRNL and assign edge weights to graphs with and without the target link, $\mathcal{N}^+$ and $\mathcal{N}^-$;
        \State Compute persistence diagrams $D^+$ and $D^-$ for $\mathcal{N}^+$ and $\mathcal{N}^-$ using edge weights as filtration values;
        \State Vectorize $D^+$ and $D^-$ into persistence images $PI^+$ and $PI^-$;
        \State If $k \neq l$, repeat the same process with the $(l,k)$-angle hop subgraph and average the two vectors;
        \State Concatenate $PI^+$ and $PI^-$ to form $PI_{(k,l)}$;
        \EndFor
        \EndFor
        \State $PIS_i = \{PI_{(k,l)} : 0 \leq l \leq k \leq H, k > 0\} = \{ PI_1, PI_2, \cdots, PI_N \}$;
        \EndFor
        \State Form training dataset $\mathcal{X} = \{ PIS_1, PIS_2, \cdots, PIS_{2n}\}$ and corresponding labels $Y=\{1,\cdots,1,0,\cdots,0\}$;
        \For{$e=1\cdots Epochs$}
        \State Sample a random batch $\{pis_1, pis_2, \cdots, pis_m \}$ from $\mathcal{X}$ and corresponding labels $\{y_1, y_2, \cdots, y_m\}$;
        \State Predict labels from classifiers as $p_i = \sum_{j=1}^N \alpha_j \Phi_j(pi_j)$, where $pis_i = (pi_1, \cdots, pi_N)$;
        \State Train the classifiers and update $\alpha$ using binary cross-entropy loss $BCE$:
        \normalsize
        \begin{center}{$\nabla_{\theta,\alpha}( \sum\limits_{pis_i \in \mathcal{X}}BCE(p_i, y_i))$};\end{center}
        \EndFor
    \end{algorithmic}
\end{algorithm}

\end{document}